\documentclass{article}
\PassOptionsToPackage{numbers, compress}{natbib}
\PassOptionsToPackage{dvipsnames,svgnames,table}{xcolor}
\usepackage[preprint]{neurips_2026}
\usepackage[utf8]{inputenc} 
\usepackage[T1]{fontenc}    
\usepackage{hyperref}       
\usepackage{url}           
\usepackage{booktabs}       
\usepackage{amsfonts}    
\usepackage{nicefrac}       
\usepackage{microtype}      
\usepackage{xcolor}         
\usepackage{subcaption}    
\usepackage[export]{adjustbox} 
\usepackage{wrapfig}       

\usepackage{authblk}

\usepackage{graphicx}
\usepackage[most]{tcolorbox}
\usepackage{fontawesome5}
\usepackage{tabularx}
\usetikzlibrary{positioning,fit}

\newcommand{\axismark}[1]{%
  \tikz[baseline=-0.45ex]{%
    \filldraw[#1, draw=#1!75!black, line width=0.3pt, rounded corners=1pt]
      (0,0) rectangle (0.7em,0.7em);%
  }%
}
\newcommand{\dchip}[2]{%
  \tikz[baseline]{%
    \node[%
      fill=#1!15, draw=#1!55!black, line width=0.35pt,%
      rounded corners=2.5pt, inner xsep=3.5pt, inner ysep=1.6pt,%
      font=\sffamily\scriptsize, anchor=base%
    ] {\strut #2};%
  }%
}

\newcommand{\modelchip}[1]{\textcolor{RoyalBlue}{\texttt{\small#1}}}
\newcommand{\primchip}[1]{\textcolor{Plum}{\texttt{\small#1}}}
\newcommand{\statbad}[1]{%
  \dchip{BrickRed}{#1}%
}
\newcommand{\statgood}[1]{%
  \dchip{OliveGreen}{#1}%
}

\definecolor{promptKw}   {HTML}{6A1B9A}
\definecolor{promptVar}  {HTML}{C2410C}
\definecolor{promptCmt}  {HTML}{6B7280}
\definecolor{UnoBlueHead}{HTML}{EAF3FF}
\definecolor{UnoBlueBand}{HTML}{F7FBFF}
\definecolor{UnoPinkHead}{HTML}{FDEAF1}
\definecolor{UnoPinkBand}{HTML}{FFF7FA}
\definecolor{UnoGreenHead}{HTML}{EAF7EC}
\definecolor{UnoGreenBand}{HTML}{F7FCF8}
\definecolor{UnoTableBest}{HTML}{E3F6E8}
\definecolor{UnoTableRule}{HTML}{A7B8C8}

\definecolor{CMathL1}{HTML}{C7D6F1}
\definecolor{CMathL2}{HTML}{D7E0F4}
\definecolor{CMathL3}{HTML}{E4EAF8}
\definecolor{CMathL4}{HTML}{F1F4FC}
\definecolor{CCodeL1}{HTML}{C8E5DD}
\definecolor{CCodeL2}{HTML}{D6EAE3}
\definecolor{CCodeL3}{HTML}{E3EFE9}
\definecolor{CCodeL4}{HTML}{EFF7F3}
\definecolor{CKnowL1}{HTML}{DDC8E5}
\definecolor{CKnowL2}{HTML}{E6D5EC}
\definecolor{CKnowL3}{HTML}{ECE2F1}
\definecolor{CKnowL4}{HTML}{F4ECF7}
\definecolor{CReadL1}{HTML}{F8D6C5}
\definecolor{CReadL2}{HTML}{FADFCE}
\definecolor{CReadL3}{HTML}{FCE7D7}
\definecolor{CReadL4}{HTML}{FDF1E8}
\definecolor{CAgL1}  {HTML}{F4CBCB}
\definecolor{CAgL2}  {HTML}{F6D5D5}
\definecolor{CAgL3}  {HTML}{F9DEDE}
\definecolor{CAgL4}  {HTML}{FCEEEE}
\definecolor{CEffL1} {HTML}{C9E0D6}
\definecolor{CEffL2} {HTML}{D5E5DB}
\definecolor{CEffL3} {HTML}{E1EBE0}
\definecolor{CEffL4} {HTML}{EEF4F0}
\definecolor{HMath}{HTML}{B8C9EA}
\definecolor{HCode}{HTML}{B8DCD2}
\definecolor{HKnow}{HTML}{D2BBE0}
\definecolor{HRead}{HTML}{F4C5AE}
\definecolor{HAg}  {HTML}{ECBABA}
\definecolor{HEff} {HTML}{B5D2C4}
\newcommand{\gainup}[1]{\,\textcolor{OliveGreen!75!black}{\scriptsize$\uparrow$#1}}
\newcommand{\gaindn}[1]{\,\textcolor{BrickRed!80!black}{\scriptsize$\downarrow$#1}}
\newcommand{\promptkw}[1]{\textcolor{promptKw}{\textbf{\texttt{#1}}}}
\newcommand{\promptvar}[1]{\textcolor{promptVar}{\texttt{\{\{#1\}\}}}}
\newcommand{\promptcmt}[1]{{\color{promptCmt}\textit{\small#1}}}

\newcommand{\PromptHead}[2]{\textbf{\textcolor{#1}{#2:}}}
\newcommand{\PromptRole}    {\PromptHead{RoyalBlue}{ROLE}}
\newcommand{\PromptContext} {\PromptHead{TealBlue}{CONTEXT}}

\newcommand{\PromptInputs}  {\PromptHead{Plum}{INPUTS}}
\newcommand{\PromptTask}    {\PromptHead{Maroon}{TASK}}
\newcommand{\PromptProc}    {\PromptHead{Maroon}{PROCEDURE}}

\newcommand{\PromptRules}   {\PromptHead{BrickRed}{HARD RULES}}
\newcommand{\PromptConstr}  {\PromptHead{BrickRed}{CONSTRAINTS}}
\newcommand{\PromptForbid}  {\PromptHead{BrickRed}{FORBIDDEN}}
\newcommand{\PromptGuide}   {\PromptHead{ForestGreen}{COST GUIDANCE}}
\newcommand{\PromptOutput}  {\PromptHead{MidnightBlue}{OUTPUT FORMAT}}

\newcounter{promptctr}

\newtcolorbox[use counter=promptctr, number format=\arabic]{promptbox}[3][\faLightbulb]{%
  enhanced jigsaw,           
  breakable,
  enforce breakable,      
  pad at break*=2mm,          
  width=\textwidth,
  arc=4mm,
  colback=white,
  colframe=NavyBlue,
  colbacktitle=NavyBlue,
  coltitle=white,
  fonttitle=\bfseries,
  drop shadow={NavyBlue!50!black},
  title={#1\ \ Prompt~\thetcbcounter:\ #2},
  label={prompt:#3},
  before skip=4pt, after skip=4pt,
  boxrule=0.8pt,
  left=8pt, right=8pt, top=6pt, bottom=6pt,
  parbox=false,
  before upper={%
    \interlinepenalty=0%
    \widowpenalty=0%
    \clubpenalty=0%
    \displaywidowpenalty=0%
    \brokenpenalty=0%
  },
}

\title{Uno-Orchestra: Parsimonious Agent Routing via Selective Delegation}

\author{
Zhiqing Cui\thanks{Project lead. \texttt{zhiqing@nuist.edu.cn}}, 
Haotong Xie\thanks{Equal contribution.}, 
Jiahao Yuan, 
Cheng Yang, 
Hanqing Wang, 
Yuxin Wu\\
Yifan Wu, 
Siru Zhong, 
Tao Yu, 
Yifu Guo, 
Siyu Zhang, 
Xinlei Yu, 
Qibing Ren, 
Usman Naseem
}

\begin{document}

\maketitle

\begin{center}
\small
\vspace{-2em}
\includegraphics[height=10pt]{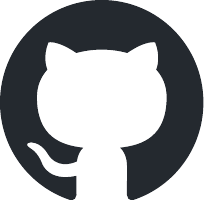} \textbf{GitHub: }
\url{https://github.com/CuiZHIQ/Uno-Orchestra} \\
\includegraphics[height=10pt]{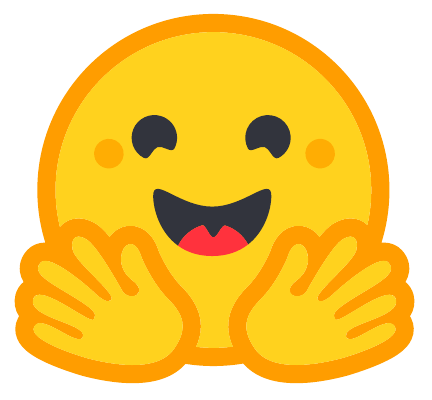}\textbf{ Dataset: }
\url{https://huggingface.co/datasets/tinaxie/Uno-Curriculum}
\end{center}
\vspace{0.5em}
\begin{abstract}
Large language model (LLM) multi-agent systems typically rely on rigid orchestration, committing either to flat per-query routing or to hand-engineered task decomposition, so decomposition depth, worker choice, and inference budget are not jointly optimized under one objective. We introduce \textsc{Uno-Orchestra}, a unified orchestration policy that selectively decomposes a task and dispatches each subtask to an admissible (model, primitive) pair, with both decisions learned together from curated RL trajectories grounded in real worker interactions. Against 22 baselines on a 13-benchmark suite spanning math, code, knowledge, long-context, and agentic tool-use, \textsc{Uno-Orchestra} reaches 77.0\% macro pass@1, roughly 16\% above the strongest workflow baseline, at roughly an order of magnitude lower per-query cost, advancing the accuracy-efficiency frontier of selective delegation.
\end{abstract}
\section{Introduction}

\begin{figure}[h!]
\centering
\includegraphics[width=0.9\linewidth]{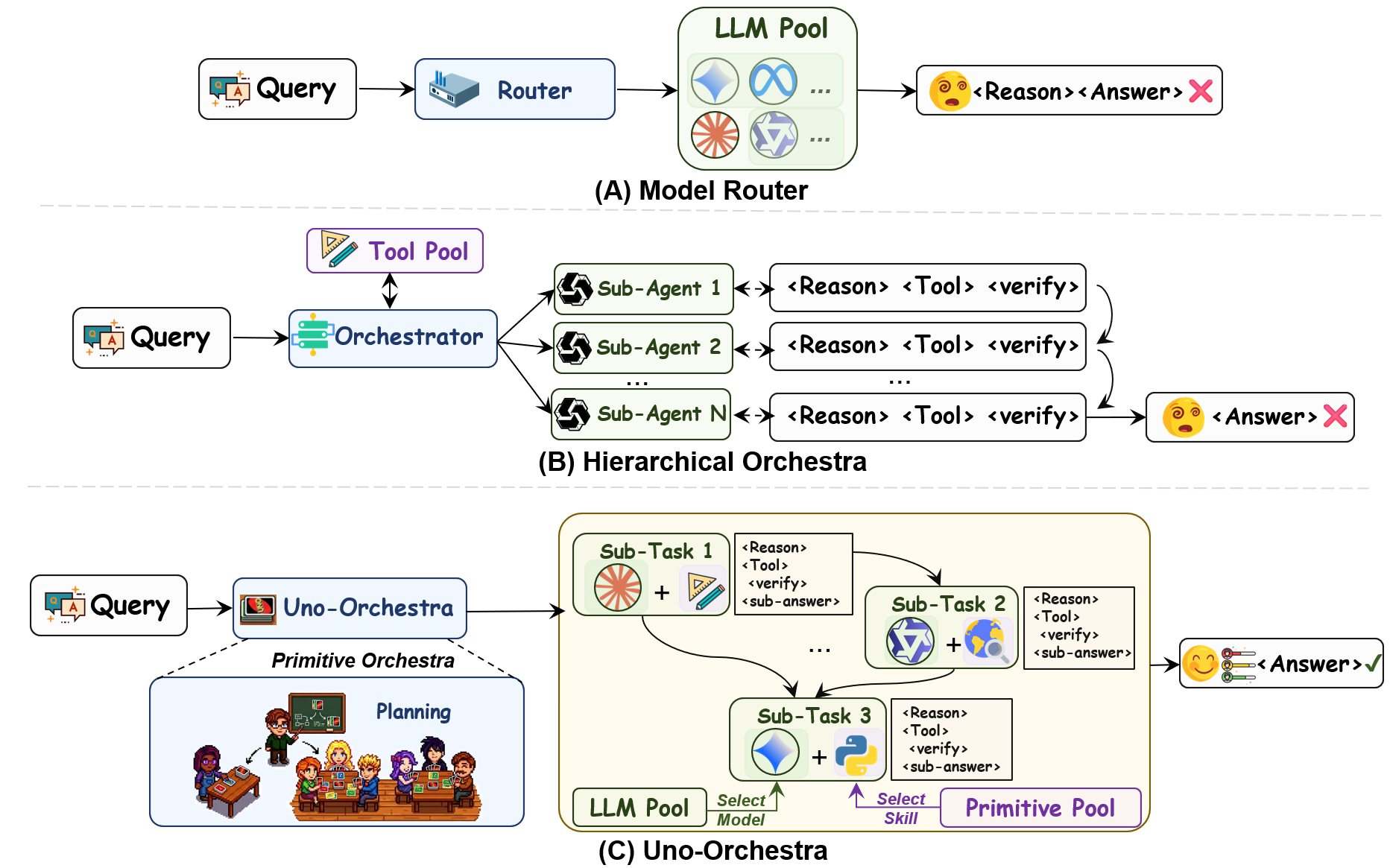}
\caption{LLM orchestration paradigms: \textbf{(A)}~model router, \textbf{(B)}~hierarchical orchestra, \textbf{(C)}~\textsc{Uno-Orchestra} (ours).}
\label{fig:teaser}
\end{figure}

Large language models have rapidly evolved from monolithic problem solvers into the building blocks of complex agentic systems that plan, retrieve and execute tools on behalf of human users \citep{ferrag2025llm, fang2025comprehensive,yuan2026decoder}. As both the capability frontier and the per-token cost of these models stretch over more than an order of magnitude \citep{chen2021evaluating}, the dominant deployment pattern of routing every query to a single model has become both wasteful and unreliable, and a growing body of work argues that capability now emerges from a learned controller that coordinates a population of heterogeneous models and tools rather than from any single model in isolation \citep{yehudai2025survey, adimulam2026orchestration}.

A general-purpose orchestrator must therefore learn \emph{when} and \emph{how} to decompose a task before any routing decision is made. Following capability taxonomies for agentic LLMs~\citep{mialon2023gaia, wang2026skillorchestra}, we optimized this requirement along six \emph{capability axes} that jointly span our training corpus and evaluation suite: atomic reasoning, compositional reasoning, knowledge retrieval, multi-hop composition, tool \& code use, and agentic \& long-context tasks. These axes demand a spectrum of orchestrator behaviours: simple queries are answered directly, while harder ones decompose into multiple coupled subtasks whose declared dependencies form a small dependency graph.

Two largely parallel research threads have grown out of this premise. The first treats orchestration as \emph{model routing}, in which a lightweight router dispatches each query to one expert: recent variants learn training-free from in-context skill profiles or generated data \citep{wang2026icl, niu2026routing, liu2026task} and end-to-end via reinforcement signals over multi-round, budgeted dispatch \citep{zhang2025router, shao2025route, okamoto2026explainable, li2026llmrouterbench}. The second treats orchestration as \emph{task decomposition over a graph of sub-agents}, in which a planner spawns specialised workers and aggregates their outputs \citep{dang2025multi, schroeder2025thread, zhuge2024gptswarm, xu2026wideseek}.

In practice, however, both threads degenerate into limited patterns. (1) \emph{Single-call routers commit at query granularity}: they pick one expert per query, miss the parallel structure of complex tasks, and the few budgeted variants \citep{qian2025xrouter, fan2026graphrag} live in a flat single-call action space that cannot trade decomposition depth against execution budget. (2) \emph{Hierarchical orchestrators decouple decomposition from dispatch}: a planner produces sub-agent calls, but the assignment of each subtask to an executor is typically hand-engineered, prompt-driven, or tied to a proprietary frontier model. As a result, the two decisions that jointly shape the quality-efficiency trade-off, whether and how deeply to decompose a task and which expert to route each subtask to, are not optimized under a unified objective.

This pattern raises the central question of our work. \emph{Can a single orchestrator learn, end to end, to decompose a task into subtasks and route every subtask to a (model, primitive) pair under an explicit cost budget?} Here a \emph{primitive} denotes the atomic routable action a worker performs on a subtask, subsuming model-internal cognitive operations, multi-step skill invocations, and external-tool calls under a single closed vocabulary. We answer affirmatively with \textsc{Uno-Orchestra}, a router-orchestrator that, given an arbitrary task, emits a plan over subtasks together with an explicit (model, primitive) routing decision for every subtask, all from a single causal-LM policy. The key design choice is to collapse decomposition and dispatch into a \emph{single shared backbone}: one causal-LM jointly emits the task decomposition and the per-subtask (model, primitive) routing within the same assistant turn, eliminating redundant context passes between separate planner and dispatcher modules.

The policy is trained in two stages. We first apply supervised fine-tuning on a verifier-gated curriculum of $61{,}201$ teacher-distilled trajectories drawn from $38$ public datasets disjoint from our evaluation suite, with code and tool subtasks executed in real sandboxes so that observations are runtime outputs rather than simulated traces. We then refine the policy on the residual hard pool with \textsc{Agentic-GRPO}, a multi-turn extension of GRPO~\citep{shao2024deepseekmath}: leveraging the orchestrator's structural cues, it augments each rollout with observation- and process-level intermediate rewards that adapt to varied decomposition patterns and resolve the long-horizon credit-assignment problem.

Our contributions are threefold:
\begin{itemize}
    \item We formulate selective delegation as a unified causal-LM policy that jointly emits subtask decomposition and per-subtask (model, primitive) routing decisions.
    \item We build a verifier-gated training pipeline and introduce Agentic-GRPO, an agent-adaptive RL objective with structured credit assignment for multi-turn orchestration.
    \item We evaluate \textsc{Uno-Orchestra} on a 13-benchmark suite and show that \textsc{Uno-Orchestra} reaches 77.0\% macro pass@1, roughly 16\% above the strongest workflow baseline, at roughly an order of magnitude lower per-query cost.
\end{itemize}

\section{Related Work}
\subsection{LLM-based Task Orchestration and Dynamic Routing}

Work on coordinating heterogeneous language models bifurcates along two dimensions: the granularity of the routing decision, and whether the routing policy is learned. At query granularity, a router selects one expert per request, an abstraction instantiated training-free through in-context model representations \citep{wang2026icl}, generated-data skill estimation \citep{niu2026routing}, and task-profile-guided synthesis for cold-start regimes \citep{liu2026task}, as well as through reinforcement learning of multi-round dispatch and aggregation \citep{zhang2025router, shao2025route}. A subset of these methods incorporates the cost of each call into the optimisation \citep{qian2025xrouter, fan2026graphrag} or pursues explainability of routing decisions \citep{okamoto2026explainable}, and the rapid growth of this family has motivated dedicated benchmarks \citep{li2026llmrouterbench}.

At subtask granularity, orchestration becomes hierarchical: a planner decomposes a task into subtasks and dispatches each to a specialised worker. Hand-designed planner-worker trees~\citep{zhang2025agentorchestra, wu2026atlas, su2025toolorchestra} fix both the decomposition and the routing policy in advance, while later work learns or evolves the sub-agent set, the topology, or the multi-agent routing itself~\citep{ruan2026aorchestra, dang2025multi, zhang2026evoroute, zhuge2024gptswarm, yue2025masrouter}; production interfaces such as Claude Code's subagent layer~\citep{anthropic2025claudecode} expose similar primitives at the API layer. Across both granularities, the decomposition and routing policies are typically optimized in isolation, and budgeted variants exist almost exclusively in the flat single-call regime.

\subsection{Reinforcement Learning for Agentic Alignment}

Reinforcement learning is the dominant paradigm for long-horizon LLM alignment, ranging from preference-based RLHF optimized with PPO \citep{schulman2017ppo, ouyang2022instructgpt} to the reward-model-free DPO objective \citep{rafailov2023dpo}. Group Relative Policy Optimisation (GRPO) \citep{shao2024deepseekmath} replaces the value network with a group-relative advantage estimator and underpins outcome-reward reasoning models such as DeepSeek-R1 \citep{guo2025deepseek}.

However, vanilla GRPO transfers poorly to multi-turn agentic rollouts where the reward arrives only at termination and the group-relative advantage is averaged over an entire trajectory. A recent line of work attacks this credit-assignment gap with finer-grained advantages: turn-level attribution~\citep{zeng2025reinforcing}, action-level inner groups~\citep{feng2025group}, tree-advantage variants that reuse partial trajectories across branches~\citep{ji2025tree, ding2025treegrpo}, and breadth-axis extensions that share rollouts across cooperating agents~\citep{xu2026wideseek}.

\section{Preliminary}
\label{sec:prelim}

We frame agentic orchestration as a multi-turn decision process under an explicit cost budget. For each query $q\sim\mathcal{Q}$, a controller iteratively plans subtasks, dispatches each to a heterogeneous worker, receives an observation, verifies progress, and emits a final answer.

\paragraph{Workers, primitives, and admissible pairs.}
Let $\mathcal{M}$ be a closed pool of frozen worker LLMs and $\mathcal{S}$ a closed vocabulary of routing primitives spanning external-tool invocations and model-internal cognitive operations. Workers cover $\mathcal{S}$ heterogeneously, so the controller selects from a sparse \emph{admissible set} $\mathcal{P}\subseteq\mathcal{M}\times\mathcal{S}$ of (model, primitive) pairs $p=(m,s)$, each carrying a token-level cost $c(p)$ that varies by orders of magnitude across $\mathcal{P}$.

\paragraph{Trajectory and policy.}
At turn $t$ the controller emits an action $a_t\sim\pi_\theta(\cdot\mid h_t)$ with history $h_t=(q,a_{<t},o_{<t})$; $a_t$ contains a plan over subtasks together with one routing pair from $\mathcal{P}$ for each subtask. The environment dispatches the routed calls and returns an observation $o_t$. A horizon-$T$ trajectory ends with a final answer $y$,
\begin{equation}
  \tau \;=\; (q,\,a_1,o_1,\,\ldots,\,a_T,o_T,\,y).
\end{equation}

\paragraph{Verifier.}
Each source provides a verifier $V$ that scores the final answer against the gold under a source-specific equivalence (symbolic for math, exact-match or F1 for QA, sandbox tests for code, schema match for tool-use), thresholded into a binary correctness signal $b=V(q,y)\in\{0,1\}$. Intermediate observations are interaction feedback, not dataset labels.

\paragraph{Objective.}
With trajectory cost $c(\tau)=\sum_{t=1}^{T}\sum_{p\in a_t}c(p)$, we train the controller to maximise
\begin{equation}
  J(\theta) \;=\;
  \mathbb{E}_{q\sim\mathcal{Q},\;\tau\sim\pi_\theta}
  \!\bigl[\, U\!\bigl(b,\, c(\tau);\,\alpha\bigr) \,\bigr],
\end{equation}
where $\alpha\in[0,1]$ trades off correctness against cost; the concrete forms of $U$ and $\pi_\theta$ are given in \S\ref{sec:uno-orchestra}.

\section{Method}
\label{sec:uno-orchestra}

\begin{figure}[!tpb]
\setlength{\abovecaptionskip}{6pt}
\centering
\includegraphics[width=\textwidth]{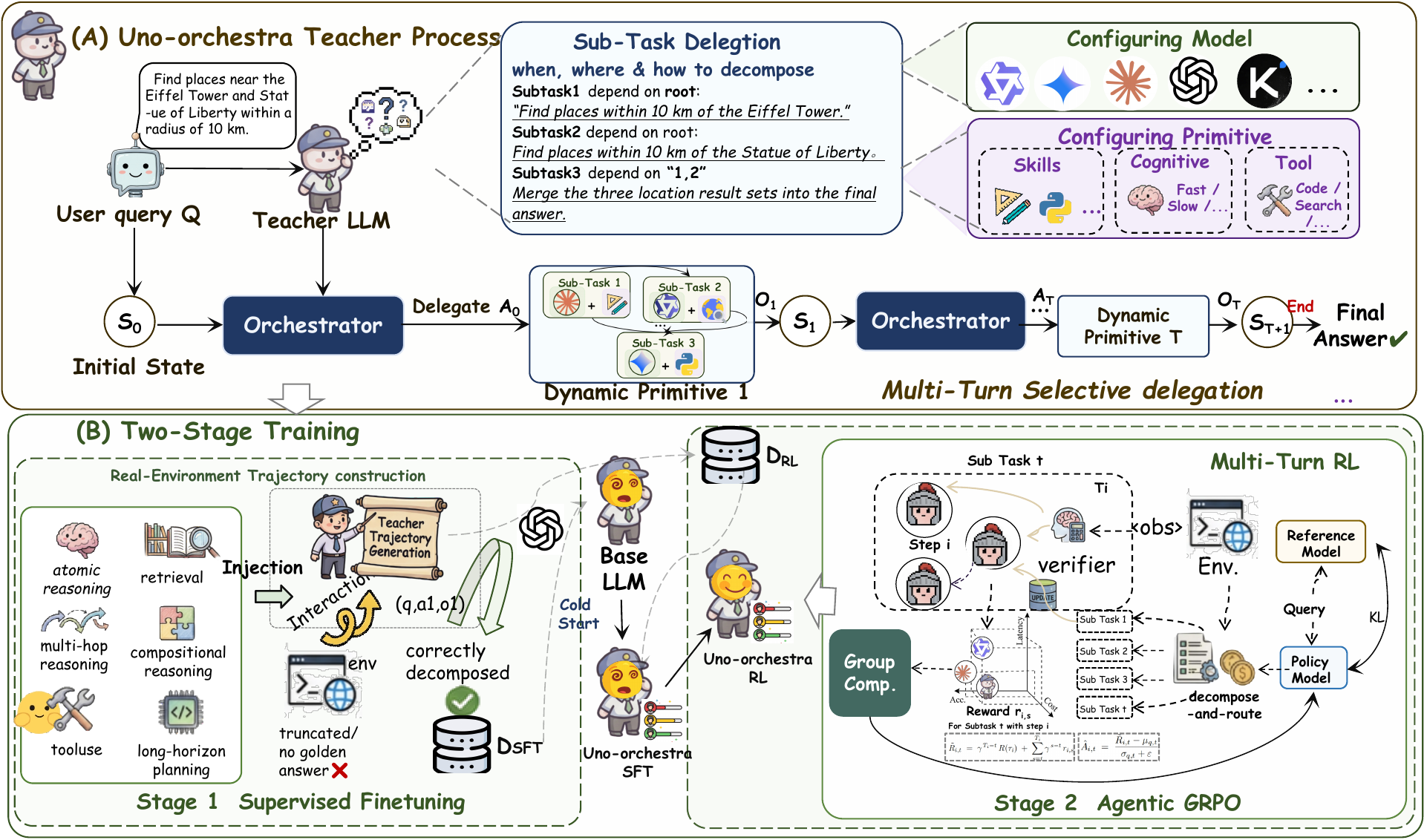}
\vspace{-1.3em}
\caption{Overview of \textsc{Uno-Orchestra}. \textbf{(A)~Multi-turn selective delegation:} at each turn the orchestrator decides \emph{when}, \emph{where}, and \emph{how} to decompose the task, configures one (model, primitive) routing pair per subtask, and dispatches the subtasks to heterogeneous workers; observations feed back as conditioning context for the next turn. \textbf{(B)~Two-stage training:} \emph{Stage~1} (SFT) distils teacher trajectories grounded in real environment interactions across our six capability axes, retaining only correctly decomposed gold-aligned trajectories into $\mathcal{D}_{\text{SFT}}$; \emph{Stage~2} (Agentic-GRPO) refines the policy on the residual hard pool $\mathcal{D}_{\text{RL}}$ using a verifier, group-relative comparison, and a KL regulariser to a frozen reference.}
\label{fig:framework}
\vspace{-1.em}
\end{figure}

We instantiate $\pi_\theta$ from \S\ref{sec:prelim} as a single causal language model. At each assistant turn it emits a plan together with one routing pair $(m,s)\in\mathcal{P}$ per subtask, where the \emph{primitive} $s$ denotes the atomic action that worker $m$ performs on the subtask, drawn from a closed vocabulary that subsumes model-internal cognitive operations (e.g.\ direct answer, chain-of-thought reasoning), multi-step skill invocations (e.g.\ document reading, code parsing), and external-tool calls (e.g.\ code execution, retrieval). Worker outputs return as observations and feed back into the policy as conditioning context for the next turn, which either replans, repairs, or emits the final answer. Training proceeds in two stages (Fig.~\ref{fig:framework}): (i)~supervised fine-tuning on a verifier-gated teacher curriculum, then (ii)~agent-adaptive reinforcement learning on the residual hard pool.

\paragraph{Verifier-gated curriculum.}
Both stages share a source-verifiable pool $\mathcal{D}_0$. Let $\pi^{(0)}$ be a cold-start router and $\pi^\star$ a strong teacher orchestrator, with verifier scores $b^{(0)}(q)=V(q,\pi^{(0)}(q))$ and $b^\star(q)=V(q,\pi^\star(q))$. We discard tasks already solved by $\pi^{(0)}$ and split the remainder by teacher outcome,
\begin{equation}
  \mathcal{D}_{\text{SFT}} \;=\; \bigl\{\bigl(q,\tau^\star(q)\bigr) : b^{(0)}\!=0,\; b^\star\!=1\bigr\},
  \qquad
  \mathcal{D}_{\text{RL}} \;=\; \bigl\{q : b^{(0)}\!=0,\; b^\star\!=0\bigr\},
\end{equation}
so $\mathcal{D}_{\text{SFT}}$ supplies behaviour-cloning targets via verifier-passing teacher trajectories $\tau^\star(q)$, and $\mathcal{D}_{\text{RL}}$ collect the residual hard pool where exploration is required.

\paragraph{Selective delegation.}
At each turn $t$, $\pi_\theta$ chooses between a \emph{direct answer} that terminates with $y$ and a \emph{decompose-and-route} action emitting a plan whose $K_t$ subtasks, together with their declared dependencies, form a small dependency graph; each subtask carries one routing pair $p_{t,k}\in\mathcal{P}$, and independent subtasks are dispatched in parallel. Plan and routing tokens are produced by the \emph{same} backbone in a single forward pass, so the joint distribution factorises along the token order:
\begin{equation}
  \pi_\theta(a_t \mid h_t)
  \;=\;
  \underbrace{\pi_\theta(\mathrm{plan}_t \mid h_t)}_{\text{decompose}}
  \;\cdot\;
  \prod_{k=1}^{K_t}
  \underbrace{\pi_\theta\bigl(p_{t,k} \mid h_t,\, \mathrm{plan}_t,\, p_{t,<k}\bigr)}_{\text{route}},
\end{equation}
the two factors being separated only by left-to-right causal masking, so the decompose-then-route structure is obtained without auxiliary heads, sub-networks, or per-stage loss weights. Selectivity is the central property: simple queries collapse to a single direct-answer turn at zero dispatch cost, while only genuinely compositional tasks pay for multi-call orchestration.

\subsection{Outcome Reward}
\label{sec:reward}

We instantiate $U$ to keep verifier correctness the dominant signal while retaining a bounded gradient on incorrect rollouts. Let $\hat{c}(\tau)\in[0,1]$ be the trajectory cost normalised by clipping $\sqrt{c(\tau)}$ against a running percentile bracket of recent rollouts. The terminal reward combines a verifier-gated cost reward with a bounded routing-shaping term,
\begin{equation}
  R(\tau) \;=\; b \,\cdot\, \bigl[\,(1-\alpha) \,+\, \alpha\,\bigl(1 - \hat{c}(\tau)\bigr)\,\bigr] \;+\; (1-b)\,\cdot\, S(\tau),
  \qquad S(\tau)\in[0,\,0.10],
\end{equation}
where $S(\tau)$ rewards schema-valid plan-and-route emissions when the verifier is unsatisfied. The cap $S\!\le\!0.10\!\ll\!1$ keeps the verifier dominant and forecloses the failure mode in which the policy harvests cheap-and-wrong cost bonuses.

\subsection{Learning to Orchestrate}
\label{sec:grpo}

\paragraph{Why intermediate credit.}
Vanilla GRPO~\citep{shao2024deepseekmath} collapses an entire rollout into a single trajectory-level return, so all assistant turns share one group-relative advantage. In multi-turn orchestration the verifier signal only arrives at termination, leaving early routing choices, repair triggers, and premature aggregation unsupervised, and the gradient flat with horizon length. Agentic-GRPO routes a portion of the terminal reward back to each turn, augmented with a bounded per-turn shaping signal, so credit is attributed at the granularity of individual orchestration decisions.

\paragraph{Turn-level return and advantage.}
For trajectory $i$ at turn $t$ with horizon $T_i$, let $r_{i,t}\in[-\eta,\eta]$ be a per-turn shaping reward (e.g.\ schema validity, repair indicator) bounded by $\eta\!\ll\!1$ so it cannot overpower the verifier. The turn-level return and its group-standardised advantage are
\begin{equation}
  \tilde{R}_{i,t}
  \;=\;
  \gamma^{\,T_i - t}\, R(\tau_i)
  \;+\;
  \sum_{s=t}^{T_i}\gamma^{\,s-t}\, r_{i,s},
  \qquad
  \hat{A}_{i,t}
  \;=\;
  \frac{\tilde{R}_{i,t} - \mu_{q,t}}{\sigma_{q,t} + \varepsilon},
\end{equation}
where $(\mu_{q,t},\sigma_{q,t})$ are the within-query mean and standard deviation of $\tilde{R}_{j,s}$ over comparable turns of the rollout group (same query, matched turn index and action type). When the rollout is a single chain, Agentic-GRPO recovers turn-level credit~\citep{zeng2025reinforcing}; when sibling turns share a prefix, it recovers branch-level credit~\citep{ji2025tree}.

\paragraph{Masked clipped objective.}
Let $\mathcal{T}_i$ be the set of token indices emitted by $\pi_\theta$ in trajectory $i$, with observation and chat-template tokens excluded, and write $t(\ell)$ for the turn containing token $\ell$. The actor loss is the clipped GRPO surrogate restricted to $\mathcal{T}_i$ plus a low-variance KL regulariser to a frozen reference $\pi_{\mathrm{ref}}$:
\begin{equation}
  \mathcal{L}(\theta)
  \;=\;
  -\,\mathbb{E}_{\tau\sim\pi_{\theta_{\mathrm{old}}}}\!\left[\,
    \sum_{i}\sum_{\ell\in\mathcal{T}_i}
    \min\!\Bigl(
      \rho_{i,\ell}\,\hat{A}_{i,t(\ell)},\;
      \mathrm{clip}(\rho_{i,\ell},\,1{-}\epsilon,\,1{+}\epsilon)\,\hat{A}_{i,t(\ell)}
    \Bigr)
  \,\right]
  \;+\; \beta\,\hat{D}_{\mathrm{KL}},
\end{equation}
with importance ratio $\rho_{i,\ell}=\pi_\theta(x_{i,\ell}\mid h_{i,\ell})\,/\,\pi_{\theta_{\mathrm{old}}}(x_{i,\ell}\mid h_{i,\ell})$ and KL estimator $\hat{D}_{\mathrm{KL}}=\rho-\log\rho-1$. The set $\mathcal{T}_i$ coincides with the SFT loss mask, so SFT and RL optimise the same router-emitted positions under different objectives.

\section{Experimental Setup}
\label{sec:exp-setup}

\paragraph{Benchmarks and metrics.}
We evaluate \textsc{Uno-Orchestra} and competing routers on $13$ benchmarks spanning five capability domains: \emph{mathematical reasoning} (MATH-500~\citep{lightman2023let}, AIME~\citep{aimo2024aime}); \emph{code \& software engineering} (HumanEval~\citep{chen2021evaluating}, MBPP~\citep{austin2021program}, LiveCodeBench~\citep{jain2024livecodebench}, SWE-bench~\citep{jimenez2023swe}); \emph{knowledge \& scientific reasoning} (MMLU~\citep{hendrycks2020measuring}, GPQA~\citep{rein2023gpqa}); \emph{reading \& long-context} (DROP~\citep{dua2019drop}, MRCR~\citep{team2024gemini}); \emph{agentic \& tool use} (GAIA~\citep{mialon2023gaia}, Terminal-Bench~\citep{team2025terminal}, ToolBench~\citep{qin2023toolllm}); plus LLMRouterBench~\citep{li2026llmrouterbench} as a routing-specialised diagnostic that is \emph{not} counted in the 13-benchmark macro-average. All evaluation sources are kept disjoint from training. We report two task-performance metrics, \textit{pass@1} (single attempt) and \textit{pass@2} (one allowed retry over independent samples).

\paragraph{Domain aggregation and RL training.}
Tab.~\ref{tab:main-results} reports macro \textit{pass@1}/\textit{pass@2} within each capability domain before efficiency columns: each domain figure is an unweighted mean over its constituent benchmarks (\textbf{Math}: MATH-500 and AIME; \textbf{Code/SE}: HumanEval, MBPP, LiveCodeBench, SWE-bench; \textbf{Know.}: MMLU and GPQA; \textbf{Read.}: DROP and MRCR; \textbf{Agentic}: GAIA, Terminal-Bench, ToolBench). The final two columns (\emph{tok} and \emph{USD/q}) report the mean context length and the billed USD per query, averaged over the same 13 benchmarks under a unified rollout harness for every competitor. RL refinement trains Agentic-GRPO~\citep{shao2024deepseekmath} on the verifier-filtered pool ($2{,}976$ questions; \S\ref{sec:grpo}) with $G{=}8$ comparison rollouts per task and at most $T_{\max}{=}8$ orchestrator turns; decoding is capped near $16{,}384$ assistant tokens ($4{,}096$ tokens of prompt retained per planner turn within that budget); optimisation uses AdamW at $10^{-6}$, PPO clipping $\epsilon{=}0.2$, KL against a frozen router reference $\beta{=}10^{-3}$, terminal cost blending $\alpha{=}0.1$ (\S\ref{sec:reward}), and collocated worker rollouts executed through vLLM. Hardware for both SFT and RL is a single $8$-GPU NVIDIA A100~80\,GB node.

\paragraph{Routing infrastructure.}
All main \textsc{Uno-Orchestra} variants share a Qwen2.5-7B-Instruct router. The \emph{main pool} is a strong heterogeneous mix of nine commercial workers spanning over two orders of magnitude in per-token price (Gemini-2.5-Flash-Lite, Gemini-2.5-Flash, Gemini-3-Flash-Preview, Gemini-3.1-Pro-Preview, Kimi-K2.5, GPT-5.3-Codex, GPT-5.4, Claude-Sonnet-4-6, Claude-Opus-4-6); for ablation we also use a \emph{Qwen2.5 scale ladder} including all different parameter sizes that removes proprietary frontier workers but preserves capability diversity by parameter scale. During RL the router observes only anonymous worker labels (\textsc{Worker}\,$1,\ldots,K$), with the mapping to backend identity hidden and resampled per episode. This forecloses brand-name shortcuts and forces the router to profile each worker through interaction and reward feedback.

\paragraph{Baselines.}
We compare against five families: \textbf{(i)} \emph{static / direct} inference (\emph{Direct Claude Opus}); \textbf{(ii)} \emph{single-round routers} (RouterDC~\citep{chen2024routerdc}, GraphRouter~\citep{fan2026graphrag}, ICL-Router~\citep{wang2026icl}, ColdStart-LLM~\citep{liu2026task}); \textbf{(iii)} \emph{multi-round / RL routers} (PromptLLM, KNN-MR, and Router-R1~\citep{zhang2025router}, R2-Reasoner~\citep{shao2026route}, AutoMix~\citep{aggarwal2023automix}, WideSeek-R1~\citep{xu2026wideseek}, xRouter~\citep{qian2025xrouter}, ATLAS~\citep{wu2026atlas} cluster/RL); \textbf{(iv)} \emph{agentic-workflow systems} (Tool Orchestra~\citep{su2025toolorchestra}, AOrchestra~\citep{ruan2026aorchestra}, AgentOrchestra~\citep{zhang2025agentorchestra}, SkillOrchestra~\citep{wang2026skillorchestra}, Puppeteer~\citep{dang2025multi}, ToolLLM~\citep{qin2023toolllm}, MasRouter~\citep{yue2025masrouter}); \textbf{(v)} \emph{ours} (five training stages summarised under Ablation Studies, \S\ref{sec:ablations}); Tab.~\ref{tab:main-results} lists the three RL checkpoints together with external baselines, and Tab.~\ref{tab:ablation-uno} reports early-stage checkpoints. All baselines share the worker pool and inference budget where applicable. Additional reporting conventions, curricula, primitives, rollout pricing rules, fuller hyper-parameters, and per-task tables are consolidated in Apps.~\ref{app:details} to \ref{app:per-benchmark}.

\section{Experimental Analysis}
\label{sec:exp-analysis}

\subsection{Main Results}
\label{sec:main-results}

\begin{figure}[!ht]
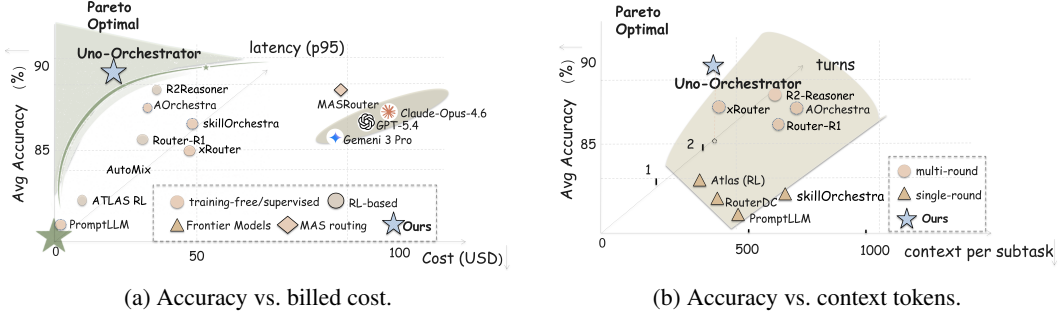

\centering
\begin{subfigure}[t]{0.48\linewidth}
\centering
\includegraphics[width=\linewidth]{Figures/uno1_02.png}
\caption{Accuracy vs.\ billed cost.}
\label{fig:acc-cost-latency-pareto}
\end{subfigure}
\hfill
\begin{subfigure}[t]{0.48\linewidth}
\centering
\includegraphics[width=\linewidth]{Figures/uno1_03.png}
\caption{Accuracy vs.\ context tokens.}
\label{fig:acc-context-turns}
\end{subfigure}
\caption{Accuracy and efficiency overview on the 13-benchmark suite. Panel~(a) plots macro pass@1 against billed cost in USD per query, with a shaded near-Pareto region and an indicative direction for tail latency (p95); panel~(b) plots the same accuracy against average context tokens per subtask, contrasting single-round and multi-round routers, and the diagonal ``turns'' cue sketches how deeper interaction shifts both accuracy and context. \textsc{Uno-Orchestra} sits on the favourable corner of both views, combining the highest macro accuracy with the smallest cost and context budgets among learned routers and agentic-workflow baselines.}
\label{fig:efficiency-tradeoffs}
\end{figure}

\begin{table}[!tb]
\centering
\caption{\textbf{Main results: five-domain macro accuracy with serving cost and context.} Macro pass@1 / pass@2 across math, code/SE, knowledge, reading, and agentic tool-use domains, together with average context tokens and per-query inference cost, comparing $22$ external baselines against three RL variants of \textsc{Uno-Orchestra}.}
\label{tab:main-results}
\footnotesize
\setlength{\tabcolsep}{3.2pt}
\renewcommand{\arraystretch}{1.06}
\arrayrulecolor{UnoTableRule}
\resizebox{\linewidth}{!}{%
\begin{tabular}{l cc cc cc cc cc cc}
\toprule
\textbf{Method}
 & \multicolumn{2}{>{\columncolor{HMath}}c}{\textbf{Math}}
 & \multicolumn{2}{>{\columncolor{HCode}}c}{\textbf{Code/SE}}
 & \multicolumn{2}{>{\columncolor{HKnow}}c}{\textbf{Know.}}
 & \multicolumn{2}{>{\columncolor{HRead}}c}{\textbf{Read.}}
 & \multicolumn{2}{>{\columncolor{HAg}}c}{\textbf{Agentic}}
 & \multicolumn{2}{>{\columncolor{HEff}}c}{\textbf{Efficiency}} \\
\cmidrule(lr){2-3} \cmidrule(lr){4-5} \cmidrule(lr){6-7} \cmidrule(lr){8-9} \cmidrule(lr){10-11} \cmidrule(lr){12-13}
 & \cellcolor{CMathL4}pass@1 & \cellcolor{CMathL4}pass@2
 & \cellcolor{CCodeL4}pass@1 & \cellcolor{CCodeL4}pass@2
 & \cellcolor{CKnowL4}pass@1 & \cellcolor{CKnowL4}pass@2
 & \cellcolor{CReadL4}pass@1 & \cellcolor{CReadL4}pass@2
 & \cellcolor{CAgL4}pass@1   & \cellcolor{CAgL4}pass@2
 & \cellcolor{CEffL4}tok     & \cellcolor{CEffL4}USD/q \\
\midrule[\heavyrulewidth]
RouterDC~\citep{chen2024routerdc} & 48.2 & 55.4 & 50.6 & 56.5 & 61.0 & 65.3 & 48.0 & 53.0 & 24.5 & 29.8 & 529 & 0.2146 \\
GraphRouter~\citep{fan2026graphrag} & 45.1 & 52.5 & 52.0 & 58.9 & 59.4 & 63.9 & 54.8 & 59.8 & 41.7 & 49.0 & 585 & 0.3084 \\
ICL-Router~\citep{wang2026icl} & 55.7 & 63.5 & 53.2 & 59.2 & 60.8 & 65.4 & 47.5 & 52.6 & 27.3 & 32.9 & 525 & 0.2100 \\
ColdStart-LLM~\citep{liu2026task} & 38.6 & 45.8 & 43.9 & 51.5 & 56.3 & 60.6 & 46.0 & 51.2 & 23.8 & 30.5 & 540 & 0.2288 \\
\midrule
PromptLLM~\citep{zhang2025router} & 40.2 & 47.2 & 51.1 & 58.3 & 57.2 & 61.9 & 54.0 & 58.9 & 40.9 & 48.2 & 747 & 0.9071 \\
R2-Reasoner~\citep{shao2026route} & 47.6 & 55.1 & 50.8 & 57.8 & 61.2 & 65.8 & 48.0 & 52.9 & 26.8 & 33.1 & 601 & 0.3564 \\
Router-R1~\citep{zhang2025router} & 38.9 & 44.5 & 12.1 & 19.4 & 53.2 & 57.3 & 39.5 & 45.0 & 13.9 & 19.7 & \cellcolor{CEffL1}\textbf{380} & \cellcolor{CEffL1}\textbf{0.0005} \\
AutoMix~\citep{aggarwal2023automix} & 34.6 & 40.8 & 46.9 & 55.0 & 53.5 & 57.2 & 42.5 & 47.8 & 22.5 & 28.3 & 1250 & 0.2955 \\
WideSeek-R1~\citep{xu2026wideseek} & 41.2 & 48.7 & 46.7 & 54.1 & 58.1 & 62.6 & 51.0 & 56.0 & 35.8 & 43.2 & 860 & 0.6485 \\
xRouter~\citep{qian2025xrouter} & 55.7 & 63.3 & 56.1 & 62.2 & \cellcolor{CKnowL4}69.4 & \cellcolor{CKnowL4}74.0 & 53.9 & 59.0 & 35.4 & 42.1 & 584 & 0.2490 \\
ATLAS (cluster)~\citep{wu2026atlas} & 56.1 & 63.8 & 49.8 & 56.9 & 60.5 & 65.0 & 49.0 & 54.2 & 29.4 & 35.9 & 662 & 0.4086 \\
ATLAS (RL)~\citep{wu2026atlas} & 52.1 & 60.7 & 51.4 & 58.3 & 62.0 & 66.8 & 50.7 & 55.5 & 32.0 & 38.8 & 736 & 0.5032 \\
\midrule
Tool Orchestra~\citep{su2025toolorchestra} & 37.6 & 42.5 & 57.9 & 62.7 & 55.3 & 59.9 & 40.6 & 40.8 & 24.6 & 29.2 & 1896 & 0.8100 \\
AOrchestra~\citep{ruan2026aorchestra} & \cellcolor{CMathL4}59.3 & \cellcolor{CMathL4}67.1 & 63.5 & 72.4 & 66.0 & 71.1 & \cellcolor{CReadL4}68.2 & \cellcolor{CReadL4}73.2 & 57.9 & \cellcolor{CAgL4}66.3 & 1600 & 0.9932 \\
AgentOrchestra~\citep{zhang2025agentorchestra} & 53.1 & 61.1 & \cellcolor{CCodeL4}71.4 & \cellcolor{CCodeL4}76.5 & 66.2 & 71.3 & 65.0 & 69.4 & \cellcolor{CAgL3}68.6 & \cellcolor{CAgL2}74.6 & 1724 & 1.2118 \\
SkillOrchestra~\citep{wang2026skillorchestra} & 53.6 & 61.1 & 55.2 & 61.2 & 63.3 & 67.9 & 50.3 & 55.3 & 35.5 & 42.0 & 533 & 0.1870 \\
Puppeteer~\citep{dang2025multi} & 47.6 & 55.1 & 51.3 & 58.3 & 61.2 & 65.8 & 48.9 & 54.0 & 29.6 & 36.4 & 694 & 0.4626 \\
ToolLLM~\citep{qin2023toolllm} & 22.7 & 28.3 & 26.3 & 33.1 & 45.5 & 49.6 & 38.4 & 43.7 & 30.1 & 34.2 & 667 & \cellcolor{CEffL3}0.1355 \\
MasRouter~\citep{yue2025masrouter} & 50.9 & 58.5 & 56.1 & 61.6 & 63.2 & 67.8 & 51.0 & 56.0 & 31.7 & 37.7 & 610 & 0.3121 \\
\midrule[\heavyrulewidth]
Uno-GRPO & \cellcolor{CMathL3}77.6 & \cellcolor{CMathL3}83.0 & \cellcolor{CCodeL3}75.4 & \cellcolor{CCodeL3}80.5 & \cellcolor{CKnowL3}79.2 & \cellcolor{CKnowL3}82.9 & \cellcolor{CReadL3}76.5 & \cellcolor{CReadL3}80.8 & \cellcolor{CAgL4}66.6 & \cellcolor{CAgL3}72.5 & \cellcolor{CEffL2}391 & \cellcolor{CEffL4}0.1610 \\
Uno-tree-GRPO & \cellcolor{CMathL2}78.5 & \cellcolor{CMathL2}84.0 & \cellcolor{CCodeL2}76.9 & \cellcolor{CCodeL2}81.7 & \cellcolor{CKnowL2}79.9 & \cellcolor{CKnowL2}83.7 & \cellcolor{CReadL2}78.2 & \cellcolor{CReadL2}82.8 & \cellcolor{CAgL2}68.9 & \cellcolor{CAgL2}74.6 & \cellcolor{CEffL4}419 & 0.1712 \\
\textbf{Uno-Orchestra} & \cellcolor{CMathL1}\textbf{79.2} & \cellcolor{CMathL1}\textbf{84.7} & \cellcolor{CCodeL1}\textbf{77.8} & \cellcolor{CCodeL1}\textbf{82.2} & \cellcolor{CKnowL1}\textbf{80.5} & \cellcolor{CKnowL1}\textbf{84.3} & \cellcolor{CReadL1}\textbf{79.7} & \cellcolor{CReadL1}\textbf{84.1} & \cellcolor{CAgL1}\textbf{70.3} & \cellcolor{CAgL1}\textbf{75.8} & \cellcolor{CEffL3}411 & \cellcolor{CEffL2}0.1011 \\
\bottomrule
\end{tabular}%
}
\arrayrulecolor{black}
\end{table}

Tab.~\ref{tab:main-pass1-perbench} reports per-benchmark \textit{pass@1} for five representative competitors and the full method, then gives the full 13-benchmark comparison across $22$ baselines and the three RL variants of \textsc{Uno-Orchestra}; Fig.~\ref{fig:efficiency-tradeoffs} visualizes the aggregate trade-offs across accuracy, cost and context that underlie those domain averages, monetary cost with an indicative p95-latency direction and context load with a turn-depth sketch. \textsc{Uno-Orchestra} is best on both macro metrics, $77.0$ pass@1 and $81.7$ pass@2, roughly 16\% and 14\% above AgentOrchestra. The gain is strongest on the agentic and reading/long-context domains, while the step from Uno-tree-GRPO to Uno-Orchestra also slightly lowers cost. This thrift is a structural property of the (model, primitive) action space: on simple queries the router collapses orchestration to a single direct dispatch to a cheap yet competent worker, paying near-zero coordination overhead, while reserving multi-call decomposition strictly for genuinely composite tasks. The same property explains why \textsc{Uno-Orchestra} undercuts every workflow baseline in cost without giving up macro accuracy, an outcome that flat routers cannot reach because they cannot decompose, and that hierarchical orchestrators cannot reach because they cannot abstain from decomposing.

\begin{table*}[!ht]
\centering
\caption{\textbf{Per-benchmark pass@1 against five representative competitors.} Columns span the 13-benchmark main suite.}
\label{tab:main-pass1-perbench}
\footnotesize
\setlength{\tabcolsep}{2.6pt}
\renewcommand{\arraystretch}{1.05}
\arrayrulecolor{UnoTableRule}
\resizebox{\textwidth}{!}{%
\begin{tabular}{l rrrrrrrrrrrrr}
\toprule
\textbf{Method} & \cellcolor{HMath}\textbf{MATH} & \cellcolor{HCode}\textbf{HE} & \cellcolor{HCode}\textbf{MBPP} & \cellcolor{HKnow}\textbf{GPQA} & \cellcolor{HKnow}\textbf{MMLU} & \cellcolor{HMath}\textbf{AIME} & \cellcolor{HAg}\textbf{GAIA} & \cellcolor{HRead}\textbf{DROP} & \cellcolor{HCode}\textbf{LCB} & \cellcolor{HRead}\textbf{MRCR} & \cellcolor{HCode}\textbf{SWE} & \cellcolor{HAg}\textbf{TBench} & \cellcolor{HAg}\textbf{ToolB} \\
\midrule[\heavyrulewidth]
ColdStart-LLM~\citep{liu2026task} & 60.9 & 72.4 & 69.8 & 35.6 & 77.1 & 16.4 & 17.2 & 59.4 & 18.7 & 32.7 & 14.6 & 12.7 & 41.6 \\
xRouter~\citep{qian2025xrouter} & \cellcolor{CMathL3}76.4 & \cellcolor{CCodeL3}88.1 & \cellcolor{CCodeL3}83.8 & \cellcolor{CKnowL2}54.7 & \cellcolor{CKnowL3}84.1 & \cellcolor{CMathL3}34.9 & \cellcolor{CAgL4}24.8 & \cellcolor{CReadL4}70.6 & \cellcolor{CCodeL3}27.6 & \cellcolor{CReadL4}37.2 & \cellcolor{CCodeL4}24.8 & \cellcolor{CAgL4}21.6 & \cellcolor{CAgL4}59.9 \\
AOrchestra~\citep{ruan2026aorchestra} & \cellcolor{CMathL2}81.7 & \cellcolor{CCodeL4}84.2 & \cellcolor{CCodeL4}80.8 & \cellcolor{CKnowL3}48.3 & \cellcolor{CKnowL4}83.7 & \cellcolor{CMathL2}36.9 & \cellcolor{CAgL3}69.4 & \cellcolor{CReadL3}73.6 & \cellcolor{CCodeL4}27.4 & \cellcolor{CReadL2}62.8 & \cellcolor{CCodeL3}61.7 & \cellcolor{CAgL3}40.6 & \cellcolor{CAgL3}63.8 \\
AgentOrchestra~\citep{zhang2025agentorchestra} & \cellcolor{CMathL4}75.1 & \cellcolor{CCodeL2}89.2 & \cellcolor{CCodeL2}85.7 & \cellcolor{CKnowL4}46.9 & \cellcolor{CKnowL2}85.6 & \cellcolor{CMathL4}31.2 & \cellcolor{CAgL1}83.4 & \cellcolor{CReadL2}75.2 & \cellcolor{CCodeL2}28.4 & \cellcolor{CReadL3}54.8 & \cellcolor{CCodeL1}82.4 & \cellcolor{CAgL2}54.2 & \cellcolor{CAgL2}68.1 \\
\midrule

\textbf{Uno-Orchestra} & \cellcolor{CMathL1}\textbf{91.9} & \cellcolor{CCodeL1}\textbf{93.1} & \cellcolor{CCodeL1}\textbf{92.4} & \cellcolor{CKnowL1}\textbf{69.2} & \cellcolor{CKnowL1}\textbf{91.8} & \cellcolor{CMathL1}\textbf{66.5} & \cellcolor{CAgL2}\textbf{82.0} & \cellcolor{CReadL1}\textbf{82.4} & \cellcolor{CCodeL1}\textbf{44.0} & \cellcolor{CReadL1}\textbf{77.0} & \cellcolor{CCodeL2}\textbf{81.8} & \cellcolor{CAgL1}\textbf{57.2} & \cellcolor{CAgL1}\textbf{71.6} \\
\midrule
\textit{Rel. gain vs best baseline} & \gainup{12.5\%} & \gainup{4.4\%} & \gainup{7.8\%} & \gainup{26.5\%} & \gainup{7.2\%} & \gainup{80.2\%} & \gaindn{1.7\%} & \gainup{9.6\%} & \gainup{54.9\%} & \gainup{22.6\%} & \gaindn{0.7\%} & \gainup{5.5\%} & \gainup{5.1\%} \\

\bottomrule
\end{tabular}%
}
\rowcolors{2}{}{}
\arrayrulecolor{black}
\end{table*}

Beyond the final score, the shape of the progression is informative. SFT delivers the first sizeable improvement by stabilising the orchestration policy; GRPO sharpens route selection; tree rollouts add a marginal increase in search coverage; and Agentic-GRPO contributes a smaller but consistent improvement by attributing credit to the turns that drive the outcome. The remaining gains therefore come from better delegation decisions rather than longer trajectories, since the controller stays near the same token budget while improving accuracy. Notably, Agentic-GRPO trims per-query cost slightly below Uno-tree-GRPO while still raising accuracy, indicating that turn-level credit removes redundant dispatches faster than it introduces new ones, so each successive RL stage allocates capacity to selecting the right worker rather than invoking additional workers.

\subsection{Generalisation across Domain Shifts}
\label{sec:domain-shift}

\begin{figure}[!ht]
\centering
\includegraphics[width=\linewidth]{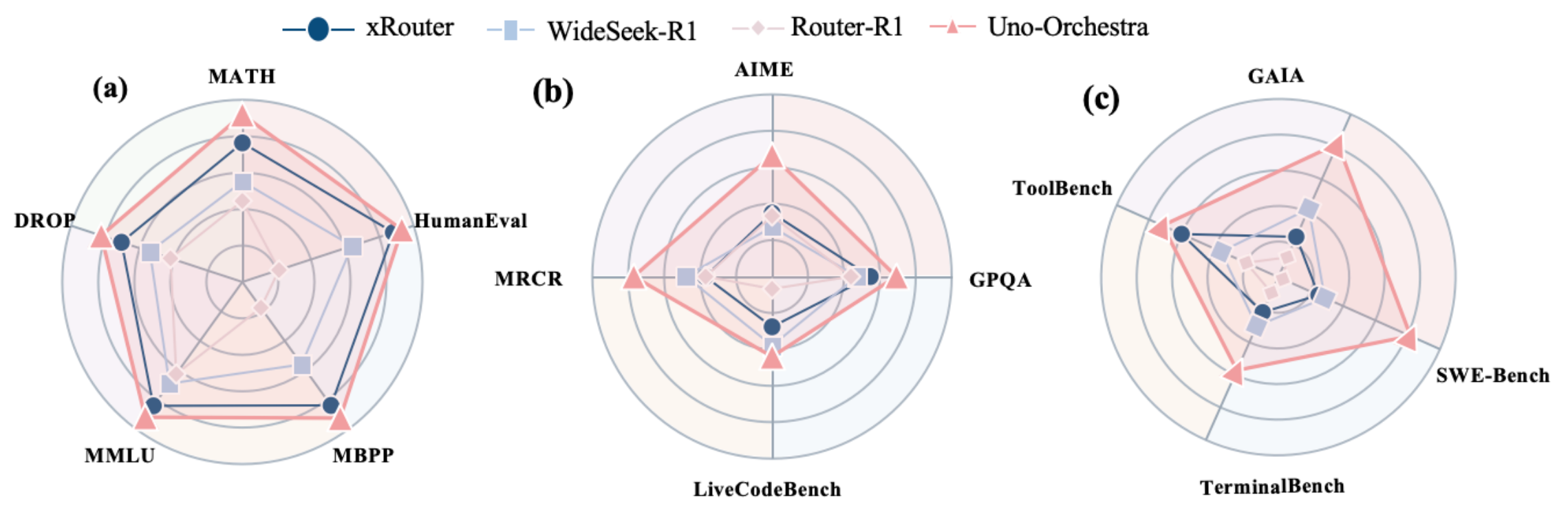}
\caption{Performance under three train-test distribution-shift regimes: (a) in-domain, (b) near-domain, and (c) out-of-domain.}
\label{fig:domain-shift}
\end{figure}

We stratify the 13-benchmark suite by its capability-axis overlap with the SFT$+$RL training pool into in-domain, near-domain, and out-of-domain regimes (Fig.~\ref{fig:domain-shift}). \textsc{Uno-Orchestra} leads in every regime and the margin grows from in-domain to out-of-domain, indicating that the learned (model, primitive) policy generalises rather than memorises the source mix.

\subsection{Ablation Studies}
\label{sec:ablations}

\textbf{Uno variants.}, 
\textbf{Uno-base} is the untrained router. \textbf{Uno-SFT} is the same router fine-tuned on the verifier-gated curriculum (\S\ref{sec:uno-orchestra}). \textbf{Uno-GRPO} adds vanilla GRPO~\citep{shao2024deepseekmath} on top of Uno-SFT with a single trajectory-level reward. \textbf{Uno-tree-GRPO} replaces single-path rollouts with prefix-sharing tree rollouts in the spirit of Tree-GRPO~\citep{ji2025tree}. \textbf{Uno-Orchestra} (the full method, \S\ref{sec:grpo}) further adds turn-level credit assignment and bounded process shaping. A compact summary of these RL objectives is given in App.~\ref{app:rl-objectives}. All Uno variants share the same router backbone, worker pool, prompt format, and inference budget.

Tab.~\ref{tab:ablation-uno} shows that the sequence Uno-base$\to$Uno-SFT$\to$Uno-GRPO$\to$Uno-tree-GRPO$\to$Uno-Orchestraimproves accuracy step by step while keeping the final controller cheaper than the tree-only variant. As an additional control, removing the blind worker protocol from Uno-GRPO lifts accuracy by less than one point per domain, yet inflates per-query cost from \$0.1610 to \$0.8163 and average context from $391$ to $548$ tokens; brand identity is therefore a shortcut the policy exploits whenever it is not foreclosed.

\begin{table}[!ht]
\centering
\caption{Stage ablation of \textsc{Uno-Orchestra}.}
\label{tab:ablation-uno}
\footnotesize
\setlength{\tabcolsep}{3.2pt}
\renewcommand{\arraystretch}{1.12}
\arrayrulecolor{UnoTableRule}
\resizebox{\linewidth}{!}{%
\begin{tabular}{l rrrr rrrrrrrr}
\toprule
& \multicolumn{4}{c}{\textbf{13-benchmark suite}} & \multicolumn{8}{c}{\textbf{rep.\ benchmarks, \textit{pass@1} (\%)}} \\
\cmidrule(lr){2-5}\cmidrule(lr){6-13}
\textbf{Variant} & \cellcolor{HMath}\textbf{pass@1} & \cellcolor{HMath}\textbf{pass@2} & \cellcolor{HEff}\textbf{USD/q} & \cellcolor{HEff}\textbf{tok} & \cellcolor{HMath}\textbf{MATH} & \cellcolor{HCode}\textbf{HE} & \cellcolor{HKnow}\textbf{MMLU} & \cellcolor{HMath}\textbf{AIME} & \cellcolor{HAg}\textbf{GAIA} & \cellcolor{HRead}\textbf{MRCR} & \cellcolor{HCode}\textbf{SWE} & \cellcolor{HAg}\textbf{ToolB} \\
\midrule[\heavyrulewidth]
Uno-base               & 48.1 & 56.1 & 0.1785 & \cellcolor{CEffL1}351 & 72.0 & 65.0 & 70.0 & 22.0 & 50.5 & 63.5 & 42.0 & 46.0 \\
Uno-SFT                & \cellcolor{CMathL4}61.3 & \cellcolor{CMathL4}68.9 & \cellcolor{CEffL3}0.1687 & \cellcolor{CEffL2}370 & \cellcolor{CMathL4}84.5 & \cellcolor{CCodeL4}79.0 & \cellcolor{CKnowL4}82.5 & \cellcolor{CMathL4}42.0 & \cellcolor{CAgL4}65.0 & \cellcolor{CReadL4}68.5 & \cellcolor{CCodeL4}56.0 & \cellcolor{CAgL4}61.5 \\
Uno-GRPO               & \cellcolor{CMathL3}74.5 & \cellcolor{CMathL3}79.5 & \cellcolor{CEffL2}0.1610 & \cellcolor{CEffL3}391 & \cellcolor{CMathL3}90.4 & \cellcolor{CCodeL3}91.8 & \cellcolor{CKnowL3}90.8 & \cellcolor{CMathL3}64.7 & \cellcolor{CAgL3}76.5 & \cellcolor{CReadL3}73.0 & \cellcolor{CCodeL3}76.0 & \cellcolor{CAgL3}68.4 \\
Uno-tree-GRPO          & \cellcolor{CMathL2}76.0 & \cellcolor{CMathL2}80.9 & \cellcolor{CEffL4}0.1712 & 419 & \cellcolor{CMathL2}91.2 & \cellcolor{CCodeL2}92.6 & \cellcolor{CKnowL2}91.3 & \cellcolor{CMathL2}65.8 & \cellcolor{CAgL2}80.6 & \cellcolor{CReadL2}75.0 & \cellcolor{CCodeL2}79.5 & \cellcolor{CAgL2}70.2 \\
\textbf{Uno-Orchestra} & \cellcolor{CMathL1}\textbf{77.0} & \cellcolor{CMathL1}\textbf{81.7} & \cellcolor{CEffL1}\textbf{0.1011} & \cellcolor{CEffL4}411 & \cellcolor{CMathL1}\textbf{91.9} & \cellcolor{CCodeL1}\textbf{93.1} & \cellcolor{CKnowL1}\textbf{91.8} & \cellcolor{CMathL1}\textbf{66.5} & \cellcolor{CAgL1}\textbf{82.0} & \cellcolor{CReadL1}\textbf{77.0} & \cellcolor{CCodeL1}\textbf{81.8} & \cellcolor{CAgL1}\textbf{71.6} \\
\bottomrule
\end{tabular}%
}
\arrayrulecolor{black}
\end{table}

\paragraph{Worker-pool diversity.}
To check that the gains do not depend on access to a specific frontier worker, we replace the main commercial pool with a Qwen2.5 scale ladder (\texttt{Qwen2.5-0.5B/1.5B/3B/7B/14B/32B/72B-Instruct}, \texttt{DeepSeek-V3} and \texttt{GPT-4o}). Tab.~\ref{tab:qwen-scale-ladder} contrasts \emph{Weaker-pools-Uno-GRPO} against a strong learned-router baseline R2-Reasoner under exactly the same pool. Despite removing every proprietary frontier worker, our orchestrator still leads R2-Reasoner across all five capability domains while incurring slightly lower context and cost. The improvement is therefore attributable to the learned (model, primitive) routing policy rather than to a single especially strong worker in the pool.

\begin{table}[!ht]
\centering
\caption{Weak-worker-pool ablation on a Qwen2.5 scale ladder. The pool removes every proprietary frontier worker but preserves capability diversity through parameter scale and a small number of public mid-size workers.}
\label{tab:qwen-scale-ladder}
\footnotesize
\setlength{\tabcolsep}{3.0pt}
\renewcommand{\arraystretch}{1.05}
\arrayrulecolor{UnoTableRule}
\resizebox{\linewidth}{!}{%
\begin{tabular}{l cc cc cc cc cc cc}
\toprule
\textbf{Method}
 & \multicolumn{2}{>{\columncolor{HMath}}c}{\textbf{Math}}
 & \multicolumn{2}{>{\columncolor{HCode}}c}{\textbf{Code/SE}}
 & \multicolumn{2}{>{\columncolor{HKnow}}c}{\textbf{Know.}}
 & \multicolumn{2}{>{\columncolor{HRead}}c}{\textbf{Read.}}
 & \multicolumn{2}{>{\columncolor{HAg}}c}{\textbf{Agentic}}
 & \multicolumn{2}{>{\columncolor{HEff}}c}{\textbf{Efficiency}} \\
\cmidrule(lr){2-3} \cmidrule(lr){4-5} \cmidrule(lr){6-7} \cmidrule(lr){8-9} \cmidrule(lr){10-11} \cmidrule(lr){12-13}
 & \cellcolor{CMathL4}p@1 & \cellcolor{CMathL4}p@2 & \cellcolor{CCodeL4}p@1 & \cellcolor{CCodeL4}p@2 & \cellcolor{CKnowL4}p@1 & \cellcolor{CKnowL4}p@2 & \cellcolor{CReadL4}p@1 & \cellcolor{CReadL4}p@2 & \cellcolor{CAgL4}p@1 & \cellcolor{CAgL4}p@2 & \cellcolor{CEffL4}tok & \cellcolor{CEffL4}USD/q \\
\midrule[\heavyrulewidth]
R2-Reasoner~\citep{shao2026route} & 47.6 & 55.1 & 50.8 & 57.8 & 61.2 & 65.8 & 48.0 & 52.9 & 26.8 & 33.1 & 601 & 0.3564 \\
\textbf{Weaker-pools-Uno-GRPO} & \textbf{48.3} & \textbf{55.8} & \textbf{51.5} & \textbf{58.5} & \textbf{61.8} & \textbf{66.4} & \textbf{48.7} & \textbf{53.6} & \textbf{27.4} & \textbf{33.8} & \textbf{589} & \textbf{0.3480} \\
\bottomrule
\end{tabular}%
}
\arrayrulecolor{black}
\end{table}

\paragraph{Router-backbone size.}
\begin{wrapfigure}{r}{0.5\linewidth}
\centering
\vspace{-0.6em}
\includegraphics[width=\linewidth]{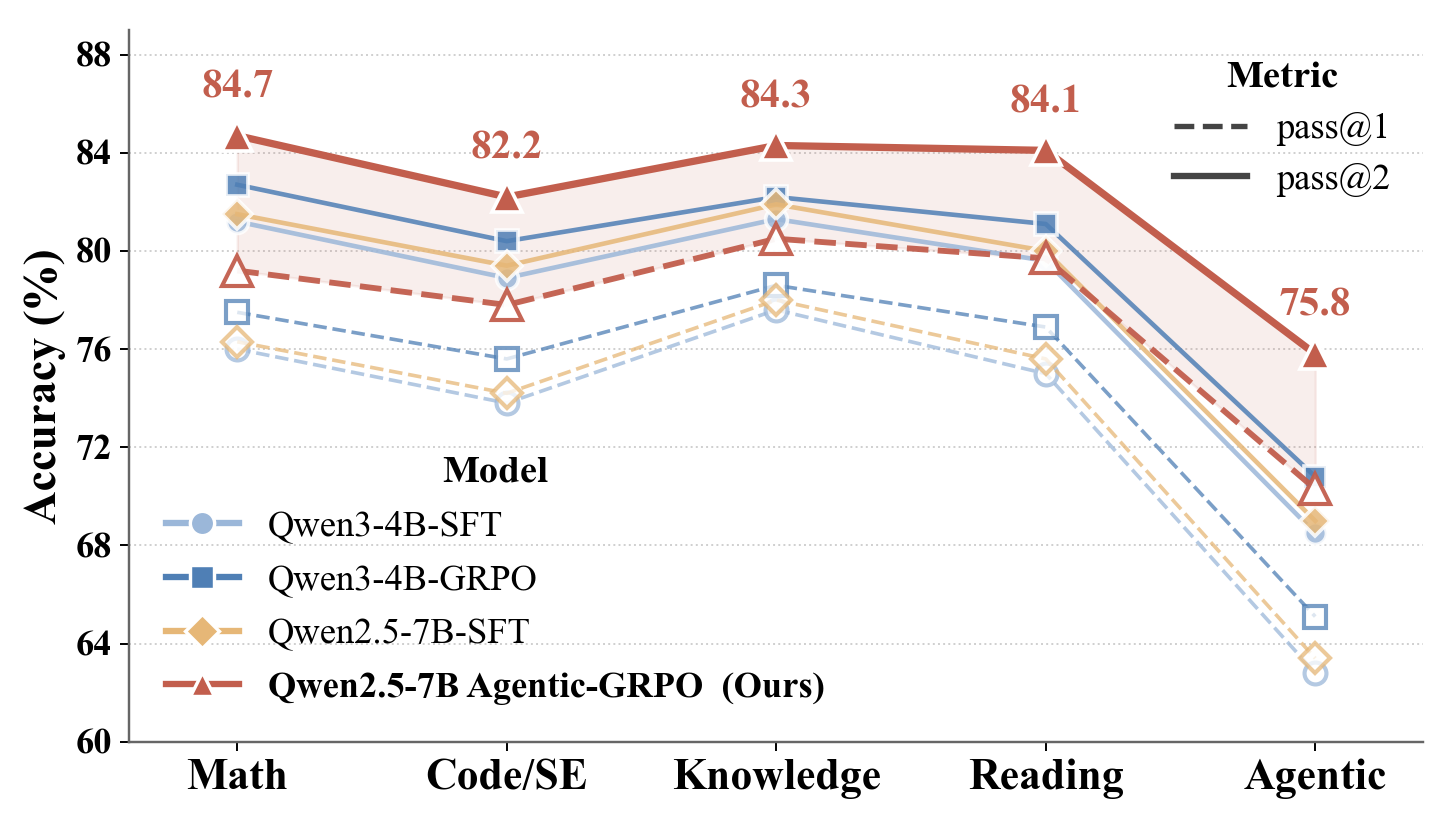}
\caption{Router-backbone comparison: pass@1 (dashed) and pass@2 (solid) across five capability domains.}
\label{fig:router-backbone}
\vspace{-1.2em}
\end{wrapfigure}
The router itself is a Qwen2.5-7B-Instruct in the main configuration. Fig.~\ref{fig:router-backbone} tests whether the policy transfers to a smaller controller by repeating the SFT and GRPO stages on a Qwen3-4B-Instruct router under the same worker pool, prompt format and inference budget. After SFT alone the 4B router nearly matches its 7B counterpart, but the RL ceiling is consistently lower across all five domains, with the largest gap on the agentic split. This suggests that controller capacity matters less for instruction following on the orchestration grammar than for credit assignment over multi-turn rollouts, which is where the additional 7B capacity is converted into accuracy.

\section{Conclusion}

We presented \textsc{Uno-Orchestra}, a unified orchestration policy that selectively decomposes a task and dispatches each subtask to an admissible (model, primitive) pair, with both decisions learned together from curated RL trajectories grounded in real worker interactions. On a 13-benchmark suite, \textsc{Uno-Orchestra} attains 77.0\% macro pass@1, roughly 16\% above the strongest workflow baseline at roughly an order of magnitude lower per-query cost, and the lead extends to weaker worker pools, smaller router backbones and out-of-domain benchmarks. Selective delegation, the ability to both decompose and abstain from decomposing under one objective, therefore offers a useful organising principle for advancing the accuracy-efficiency frontier of cost-aware multi-agent systems.

\section{Authors}
\begin{center}
\begin{tabular}{cc}
Zhiqing Cui (\textit{NUIST}) 
& Haotong Xie (\textit{SUFE}) \\[4pt]

Jiahao Yuan (\textit{ECNU}) 
& Cheng Yang (\textit{HDU}) \\[4pt]

Hanqing Wang (\textit{HKUST--GZ}) 
& Yuxin Wu (\textit{RUC}) \\[4pt]

Yifan Wu (\textit{Ramus}) 
& Siru Zhong (\textit{HKUST--GZ}) \\[4pt]

Tao Yu (\textit{CAS--ICT}) 
& Yifu Guo (\textit{SYSU}) \\[4pt]

Siyu Zhang (\textit{UC San Diego}) 
& Xinlei Yu (\textit{NUS}) \\[4pt]

Qibing Ren (\textit{SJTU}) 
& Usman Naseem (\textit{Macquarie University})
\end{tabular}
\end{center}

\bibliographystyle{plainnat}
\bibliography{main}

\appendix

\raggedbottom
\section{Experimental Details}
\label{app:details}

\subsection{Evaluation reporting conventions}
\label{app:reporting-conventions}

The main suite comprises the thirteen public benchmarks enumerated in \S\ref{sec:exp-setup}. LLMRouterBench is evaluated in parallel for routing diagnostics (App.~\ref{app:per-benchmark}) but excluded from thirteen-benchmark means. Tab.~\ref{tab:main-results} domain columns report unweighted averages of constituent tasks:

\textit{Math:} MATH-500 and AIME.

\textit{Code/SE:} HumanEval, MBPP, LiveCodeBench (LCB), and SWE-bench Verified.

\textit{Know.:} MMLU and GPQA.

\textit{Read.:} DROP and MRCR.

\textit{Agentic:} GAIA, Terminal-Bench, and ToolBench.

\textit{Efficiency columns.} Average total context (\emph{tok}) and billed dollars per assessed query (\emph{USD/q}) use published list-price token accounting common to every method and uniformly average over the thirteen primary benchmarks inside the fixed rollout harness.

\textit{Routing diagnostic.} The LLMRouterBench column complements the thirteen-task grids but probes router-only discrimination rather than orchestrated executions, so it does not enter the domain aggregates above.

\subsection{Reinforcement learning and compute stack}
\label{app:details-rl}

Agentic-GRPO fine-tunes the router on the verifier-filtered pool of $2{,}976$ RL tasks produced by curriculum promotion (\S\ref{sec:reward}, App.~\ref{app:curriculum}). Each optimisation step contrasts $G{=}8$ independent rollouts for every question subject to $T_{\max}{=}8$ orchestrator turns, decoding budget up to $L{=}16{,}384$ assistant tokens alongside the rollout-server cap, with prompts clipped to $4{,}096$ tokens before assistant generation whenever longer prefixes would overrun the rollout budget.

Advantages aggregate per (query, turn) cohort from verifier-gated trajectory scores (\S\ref{sec:grpo}); updates use AdamW at $1{\times}10^{-6}$, clipped PPO with $\epsilon{=}0.2$, KL regularisation coefficient $\beta{=}10^{-3}$ against an SFT-frozen router reference (no entropy bonus), and terminal cost blending coefficient $\alpha{=}0.1$ whose denominator tracks rolling percentiles of $\sqrt{c(\tau)}$ over the most recent $1{,}000$ completed RL episodes (\S\ref{sec:reward}). Worker calls multiplex through neighbouring vLLM~0.11 replicas with scheduler batch caps of $24{,}000$ tokens per GPU, yielding roughly $340$\,seconds per optimisation step under the eager configuration recorded in Appendix~\ref{app:training-setup}.

\textbf{Hardware and stage synopsis.},  Stage~SFT and Stage~RL occupy one host with eight NVIDIA A100 $80$\,GB accelerators wrapping the identical Qwen2.5-7B-Instruct router: SFT uses DeepSpeed ZeRO-3~\citep{rajbhandari2020zero}; RL uses FSDP ZeRO-3 with parameter and optimiser offload, with indicative wall-clock costs of roughly $6$\,hours versus $25$\,hours (App.~\ref{app:training-setup}). The SFT stage trains $61{,}201$ distilled trajectories for two cosine epochs starting after $100$ warmup steps at learning rate $2{\times}10^{-5}$, bf16 mixed precision and sequence packing up to $16{,}384$ tokens while gradients flow only on assistant completions.

Primitives, worker anonymisation protocols, verbatim schedules, rollout throughput knobs, KL estimator choice and per-scoreboard breakdowns reside in Appendix~\ref{app:training-setup}, \ref{app:skill-pool} and~\ref{app:worker-pool}; spreadsheet exports underpinning aggregates appear in Appendix~\ref{app:extended-results}; full per-task scoreboards occupy Appendix~\ref{app:per-benchmark}.

\subsection{Dataset}
\label{app:datasets}

The training corpus and the evaluation suite together draw on roughly thirty public datasets, partitioned into six \emph{capability axes} that mirror the orchestration patterns the router must master. Atomic-reasoning sources (GSM8K~\citep{cobbe2021training}, ARC~\citep{clark2018think}, MMLU~\citep{hendrycks2020measuring}, CommonsenseQA~\citep{talmor2019commonsenseqa}, PIQA~\citep{bisk2020piqa}, Social-IQa~\citep{sap2019social}, WinoGrande~\citep{sakaguchi2021winogrande}) test single-step factual or arithmetic recall and teach the router \emph{when not to decompose}. Compositional-reasoning sources (NuminaMath-CoT~\citep{li2024numinamath}, Hendrycks-MATH~\citep{hendrycks2021measuring}, MATH-500~\citep{lightman2023let}, AIME~\citep{aimo2024aime}, BBH~\citep{suzgun2023challenging}, LogiQA-v2~\citep{liu2023logiqa}, FOLIO~\citep{han2024folio}, GPQA~\citep{rein2023gpqa}) demand multi-step symbolic chains. Knowledge-retrieval sources (DROP~\citep{dua2019drop}, HotpotQA~\citep{yang2018hotpotqa}, TriviaQA~\citep{joshi2017triviaqa}, StrategyQA~\citep{geva2021did}, NQ-Open~\citep{kwiatkowski2019natural}, WebQuestions~\citep{berant2013semantic}, SciQ~\citep{welbl2017crowdsourcing}) require single- or parallel-evidence lookups. Multi-hop composition (MuSiQue~\citep{trivedi2022musique}, 2WikiMultihopQA~\citep{ho2020constructing}) exercises deep sequential dependencies between sub-queries. The \emph{tool \& code} bucket (TACO~\citep{li2023taco}, Codeforces-CoTs~\citep{openr1codeforces2025}, HumanEval~\citep{chen2021evaluating}, MBPP~\citep{austin2021program}, LiveCodeBench~\citep{jain2024livecodebench}, ToolACE~\citep{liu2024toolace}, ToolBench~\citep{qin2023toolllm}) covers sandbox-grounded operations such as competitive-programming and structured API calls. Finally, the \emph{agentic \& long-context} bucket (SWE-bench Verified~\citep{jimenez2023swe}, Terminal-Bench~\citep{team2025terminal}, GAIA~\citep{mialon2023gaia}, QuALITY~\citep{pang2022quality}, MRCR~\citep{team2024gemini}) measures full-task completion under realistic latency and context budgets. Tab.~\ref{tab:capability-datasets} groups every source by axis.

\begin{table}[ht]
\centering
\caption{Data sources organised by capability axis.}
\label{tab:capability-datasets}
\renewcommand{\arraystretch}{1.55}
\setlength{\tabcolsep}{6pt}
\begin{tabularx}{\textwidth}{@{}>{\raggedright\arraybackslash}p{4.0cm}@{\hspace{8pt}}X@{}}
\toprule
\textbf{Capability axis} & \textbf{Data sources} \\
\midrule
\axismark{RoyalBlue}\,\textbf{Atomic reasoning} &
\dchip{RoyalBlue}{GSM8K} \dchip{RoyalBlue}{ARC} \dchip{RoyalBlue}{MMLU} \dchip{RoyalBlue}{CommonsenseQA} \dchip{RoyalBlue}{PIQA} \dchip{RoyalBlue}{Social-IQa} \dchip{RoyalBlue}{WinoGrande} \\
\axismark{Plum}\,\textbf{Compositional reasoning} &
\dchip{Plum}{NuminaMath-CoT} \dchip{Plum}{Hendrycks-MATH} \dchip{Plum}{MATH-500} \dchip{Plum}{AIME} \dchip{Plum}{BBH} \dchip{Plum}{LogiQA-v2} \dchip{Plum}{FOLIO} \dchip{Plum}{GPQA} \\
\axismark{SeaGreen}\,\textbf{Knowledge retrieval} &
\dchip{SeaGreen}{DROP} \dchip{SeaGreen}{HotpotQA} \dchip{SeaGreen}{TriviaQA} \dchip{SeaGreen}{StrategyQA} \dchip{SeaGreen}{NQ-Open} \dchip{SeaGreen}{WebQuestions} \dchip{SeaGreen}{SciQ} \\
\axismark{Cerulean}\,\textbf{Multi-hop composition} &
\dchip{Cerulean}{MuSiQue} \dchip{Cerulean}{2WikiMultihopQA} \\
\axismark{BurntOrange}\,\textbf{Tool \& Code use} &
\dchip{BurntOrange}{TACO} \dchip{BurntOrange}{Codeforces-CoTs} \dchip{BurntOrange}{HumanEval} \dchip{BurntOrange}{MBPP} \dchip{BurntOrange}{LiveCodeBench} \dchip{BurntOrange}{ToolACE} \dchip{BurntOrange}{ToolBench} \\
\axismark{BrickRed}\,\textbf{Agentic \& long-context} &
\dchip{BrickRed}{SWE-bench Verified} \dchip{BrickRed}{Terminal-Bench} \dchip{BrickRed}{GAIA} \dchip{BrickRed}{QuALITY} \dchip{BrickRed}{MRCR} \\
\bottomrule
\end{tabularx}
\end{table}

\subsection{Verifier-Gated Curriculum}
\label{app:curriculum}

The pipeline that produces $\mathcal{D}_{\text{SFT}}$ and $\mathcal{D}_{\text{RL}}$ runs in five phases over a balanced raw pool of ${\sim}10\,$k tasks, drawn under per-source quotas across the capability axes of Tab.~\ref{tab:capability-datasets}.

\textbf{Verifier-gated split.}, 
Each task is probed first by the cold-start router $\pi^{(0)}$ and then by a strong teacher orchestrator $\pi^\star$, both at pass@3. Tasks already solved by $\pi^{(0)}$ are discarded as offering no learning signal; tasks with $b^\star{=}1$ enter $\mathcal{D}_{\text{SFT}}$ together with the verifier-passing teacher trajectory $\tau^\star$; tasks with $b^\star{=}0$ enter $\mathcal{D}_{\text{RL}}$. Trajectories polluted by infrastructure artefacts (API timeouts, malformed responses) are dropped.

\textbf{Failure-driven prompt repair.}, 
Failed teacher trajectories are classified by GPT-4o into four root causes (\emph{information loss}, \emph{premature aggregation}, \emph{format mismatch}, \emph{delegation scope}); each high-frequency category yields one task-agnostic constraint added to $\pi^\star$'s system prompt. Three rounds suffice: residual failures are routing-policy issues that no prompt patch can close, and the remaining tasks are therefore allocated to the RL training pool $\mathcal{D}_{\text{RL}}$ to facilitate exploration.

\textbf{Augmentation and fallback cascade.}, 
SFT questions receive two extra teacher rollouts at $\mathrm{temp}\in\{0.5,\,1.0\}$, RL-pool questions three at $\{0.3,\,0.7,\,1.0\}$; only verifier-passing rollouts survive, with the gold answer doubling as a consistency gate. RL-pool questions where the primary teacher failed are then retried under a stronger cascade (Gemini-2.5-Pro $\to$ Claude-Sonnet-4-6 $\to$ GPT-5.4); a successful rollout promotes the question from $\mathcal{D}_{\text{RL}}$ to $\mathcal{D}_{\text{SFT}}$, shrinking the RL pool from $4{,}549$ to $2{,}976$. After all phases, $|\mathcal{D}_{\text{SFT}}|=61{,}201$ and $|\mathcal{D}_{\text{RL}}|=2{,}976$; every SFT row carries provenance fields \texttt{teacher} and \texttt{distillation\_pass}$\,\in\,\{\textsf{primary},\,\textsf{augmentation},\,\textsf{fallback}\}$.

\subsection{Worker Model Pool}
\label{app:worker-pool}

The closed-vocabulary worker pool comprises nine frozen large language models drawn from four providers and spanning more than two orders of magnitude in per-token inference cost. From Google we use \texttt{gemini-2.5-flash-lite}~\citep{gemini2025gemini25}, \texttt{gemini-2.5-flash}~\citep{gemini2025gemini25}, \texttt{gemini-3-flash-preview}~\citep{gemini2025gemini3} and \texttt{gemini-3.1-pro-preview}~\citep{gemini2025gemini3}; from Moonshot, \texttt{kimi-k2.5}~\citep{moonshot2026kimik25}; from OpenAI, \texttt{gpt-5.3-codex} and \texttt{gpt-5.4}; and from Anthropic, \texttt{claude-sonnet-4-6} and \texttt{claude-opus-4-6}. Every worker is exposed only through an OpenAI-compatible chat endpoint and treated by the router as a stateless oracle, so the policy gradient is confined to the router's own tokens, and the same rollout code can target any provider gateway.

\subsection{Routing Primitive Pool}
\label{app:skill-pool}

Every \promptkw{<route>} tag commits the router to a single primitive drawn from the closed vocabulary $\mathcal{S}$, organised into clusters that share routing semantics: \emph{answer \& reason} (no external call), \emph{retrieve} (information lookup), \emph{skills} (multi-step extraction, parsing, or verification over the worker's parametric knowledge or an attached input), \emph{execute} (sandboxed code or API call) and \emph{symbolic} (computer-algebra computation). Tab.~\ref{tab:skill-pool} lists samples of each primitive alongside its functional contract.

\begin{table}[ht]
\centering
\caption{Routing primitives in $\mathcal{S}$, grouped by semantic cluster.}
\label{tab:skill-pool}
\small
\renewcommand{\arraystretch}{1.15}
\setlength{\tabcolsep}{5pt}
\begin{tabularx}{0.95\textwidth}{@{}lX@{}}
\toprule
\textbf{Primitive} & \textbf{Functional contract} \\
\midrule
\multicolumn{2}{@{}l}{\textbf{Answer \& reason}} \\
, \texttt{direct\_answer}    & Solve the subtask directly without invoking any tool or further routing. \\
, \texttt{reason}            & Produce an explicit chain-of-thought before committing to an answer. \\
\addlinespace[2pt]
\multicolumn{2}{@{}l}{\textbf{Retrieve}} \\
, \texttt{web\_search}       & Issue search-engine queries and return ranked snippets with provenance URLs. \\
, \texttt{database\_query}   & Execute a structured query against a tabular or graph knowledge base. \\
, \texttt{fact\_check}       & Verify a single factual claim against an authoritative source and return a verdict. \\
\addlinespace[2pt]
\multicolumn{2}{@{}l}{\textbf{Skills}} \\
, \texttt{read\_document}    & Read a long document and return targeted excerpts or a faithful summary. \\
, \texttt{read\_code}        & Parse a source-code file and reason about its behaviour or structure. \\
, \texttt{extract\_field}    & Return one or more named fields from a structured input payload. \\
, \texttt{parse\_structured} & Convert free-form text into a typed object such as JSON or a record. \\
\addlinespace[2pt]
\multicolumn{2}{@{}l}{\textbf{Execute}} \\
, \texttt{execute\_python}   & Emit a self-contained Python program; the sandbox returns stdout and stderr. \\
, \texttt{execute\_shell}    & Emit a single shell command; the sandbox returns the captured terminal output. \\
, \texttt{call\_api}         & Issue a typed function call to an external HTTPS service per a declared schema. \\
\addlinespace[2pt]
\multicolumn{2}{@{}l}{\textbf{Symbolic}} \\
, \texttt{symbolic\_math}    & Invoke a computer-algebra backend for exact algebraic manipulation. \\
\bottomrule
\end{tabularx}
\end{table}

\subsection{Subtask DAG and Action Schema}
\label{app:dag-schema}

The DAG shape arises directly from the dependency relations between subtasks: independent subtasks fan out and dispatch in parallel, while subtasks that consume an upstream observation contribute sequential edges, so a purely chain-shaped trajectory is recovered as a degenerate special case in which every subtask depends on its predecessor.

The plan $P_t$ emitted at each assistant turn is a directed acyclic graph $P_t=(\mathcal{V}_t,\mathcal{E}_t)$ over a set of subtask nodes. Every node $k\in\mathcal{V}_t$ carries a free-form natural-language description, an admissible routing pair $p_{t,k}=(m_{t,k},s_{t,k})\in\mathcal{P}$, and a (possibly empty) set of dependency edges $\{(j,k):j\prec k\}\subseteq\mathcal{E}_t$ specifying that $k$ must consume the observation produced by $j$. Acyclicity is enforced by the trajectory grammar (Prompt~\ref{prompt:schema}): each \texttt{<route>} chip carries an integer \texttt{id} strictly larger than every \texttt{depends\_on} entry it lists, so a valid topological order coincides with the emission order of \texttt{<route>} chips inside the same assistant turn. At dispatch time, leaves with all dependencies resolved are sent to their workers in parallel, observations are spliced back as \texttt{<obs>} blocks, and the cycle repeats until $\mathcal{V}_t$ is exhausted. Hence the action $a_t=(P_t,\{p_{t,k}\}_{k=1}^{K_t})$ encodes both the dependency structure and every per-node routing decision in a single contiguous stretch of policy tokens between two \texttt{<obs>} blocks. The same XML grammar is used for SFT supervision and RL rollouts, so the causal factorisation in \S\ref{sec:uno-orchestra} is materialised by the token order without any architectural change.

\subsection{Training Setup}
\label{app:training-setup}

\textbf{Hardware.}, 
All training runs use a single node with $8{\times}$NVIDIA A100 80\,GB GPUs. Both stages fine-tune the same Qwen2.5-7B-Instruct backbone in full precision: SFT runs under DeepSpeed ZeRO-3 \citep{rajbhandari2020zero} and finishes in roughly $6$ wall-clock hours; RL runs under FSDP ZeRO-3 with parameter and optimizer offload and finishes in roughly $25$ wall-clock hours.

\textbf{SFT hyper-parameters.}, 
Trained on the 61201 teacher trajectories for $2$ epochs at learning rate $2{\times}10^{-5}$ with cosine schedule and $100$-step warmup; bf16 mixed precision, sequence packing on, cutoff $16{,}384$ tokens, effective batch size $128$. All \texttt{observation} tokens are loss-masked; only \texttt{assistant} tokens carry gradient.

\textbf{RL hyper-parameters and throughput.}, 
Agentic-GRPO is run over the $2{,}976$ RL-pool questions. Group size $G{=}8$, max assistant turns $T_{\max}{=}8$, response budget $L{=}16{,}384$ (prompt cap $4{,}096$); AdamW learning rate $1{\times}10^{-6}$; PPO clip $\epsilon{=}0.2$; low-variance KL added to the loss with $\beta{=}10^{-3}$, no entropy bonus; cost blend $\alpha{=}0.1$ with rolling-percentile $\sqrt{c}$-normalisation over a $1{,}000$-episode buffer. Rollout server runs vLLM 0.11 and dynamic batching capped at $24{,}000$ tokens per GPU. Reference single-node throughput: ${\sim}340$\,s/step under eager mode.

\clearpage
\section{Reinforcement-Learning Objectives}
\label{app:rl-objectives}

This appendix gives a compact view of the reinforcement-learning objectives used in the ablations and in the final \textsc{Uno-Orchestra} policy. All variants share the same outer loop: for a task $x$, the router samples a small comparison set of rollouts, executes the selected workers or tools, scores each trajectory, and updates only the router tokens. Let $\pi_\theta$ be the trainable policy, $\pi_o$ the behaviour policy used for rollout collection, and $\pi_r$ the frozen reference policy for KL regularisation.

\paragraph{GRPO and Uno-GRPO.}
Uno-GRPO is the terminal-credit baseline. For a group of $G$ independent rollouts $\tau^1,\ldots,\tau^G$, each rollout receives one terminal score
\[
R_g = V_g - \alpha C_g,
\]
where $V_g$ is the verifier or task-success score and $C_g$ is the normalized resource term. Scores are compared only within the $G$ rollouts of the same task,
\[
A_g = \frac{R_g-\bar{R}}{\sigma_R+\epsilon}.
\]
The router is then optimized with the clipped ratio
\[
r_t(\theta)=\frac{\pi_\theta(a_t\mid s_t)}{\pi_o(a_t\mid s_t)}
\]
and objective
\[
\mathcal{L}_{G}(\theta)
=
-\mathbb{E}\!\left[
\min\!\left(r_t A_g,\,
\mathrm{clip}(r_t,1-\epsilon,1+\epsilon)A_g\right)
-\beta\,\mathrm{KL}(\pi_\theta\Vert\pi_r)
\right].
\]
This baseline is useful because it isolates what terminal reward alone can teach. It improves route selection after SFT, but every turn in the same trajectory receives the same advantage, so early decomposition errors and late aggregation errors are not separated.

\paragraph{Tree-GRPO and Uno-tree-GRPO.}
Uno-tree-GRPO changes the sampling geometry rather than the reward definition. Instead of drawing $G$ fully independent rollouts, the router first samples a shared prefix and then branches at selected decision points, producing sibling continuations under the same partial context. If $b$ indexes a sibling branch, the branch return is normalized within its local sibling set,
\[
T_b=\frac{R_b-\bar{R}_B}{\sigma_B+\epsilon}.
\]
Tokens before the branch receive the prefix-level signal, while tokens after the branch receive the sibling advantage $T_b$ through the same clipped objective as above. This makes the comparison sharper than vanilla GRPO: two continuations can be judged after agreeing on the same decomposition prefix, so the update can reward a better repair, worker choice or aggregation decision without conflating it with unrelated early-token variation. Its limitation is that tree expansion increases rollout context, which is why the final method keeps the useful branch-level credit but adds a more targeted process signal.

\paragraph{MT-GRPO.}
MT-GRPO keeps the same clipped update but adds turn-level credit for multi-turn traces. A turn score $U_t$ is computed from the downstream outcome of the decision at turn $t$, and a group-normalized turn advantage is formed as
\[
B_t=\frac{U_t-\bar{U}}{\sigma_U+\epsilon}.
\]
The update advantage becomes
\[
\hat{A}_t = A_g+\rho B_t,
\]
where $\rho$ controls how much local turn credit is mixed into the terminal trajectory signal. This objective is better matched to agentic routing than vanilla GRPO because a repair turn, a premature final answer, and a correct worker dispatch can receive different credit even when they occur in the same rollout.

\paragraph{GiGPO.}
GiGPO generalizes the local-credit idea by grouping actions that start from semantically equivalent anchor states. For an anchor group $C$, each action receives a local outcome estimate $Q_t$, and the micro advantage is
\[
M_t=\frac{Q_t-\bar{Q}_C}{\sigma_Q+\epsilon}.
\]
The final advantage is
\[
\tilde{A}_t = A_g+\eta M_t.
\]
The distinction from MT-GRPO is that the comparison set is not merely ``same task and same turn''. Instead, GiGPO compares routing actions that face the same local state, even if they appear in different rollouts or at different depths. This makes it attractive for long-horizon agents, but it also depends on reliable anchor construction.

\paragraph{Agentic-GRPO / Uno-RL.}
The final objective used by \textsc{Uno-Orchestra} keeps the group-relative update above but makes the local signal specific to the router's structured trajectory. We denote the strongest blind-pool variant as Uno-RL; in the main paper this corresponds to the full Agentic-GRPO checkpoint. For each turn, the policy records the emitted plan, routes, observations and verification result, then builds a shaped score
\[
S_t = w_t R_g+\gamma V_t-\alpha C_t,
\]
where $w_t$ allocates part of the terminal score to the responsible turn, $V_t$ rewards verifier-confirmed progress, and $C_t$ discourages unnecessary delegation. The normalized local term
\[
D_t=\frac{S_t-\bar{S}}{\sigma_S+\epsilon}
\]
is combined with the trajectory advantage,
\[
A_t^{\star}=A_g+D_t.
\]
This is the version best aligned with \textsc{Uno-Orchestra}: lazy answers, one-shot parallel plans, continuation chains and repair turns all remain in the same causal-LM action space, while the reward distinguishes whether a turn improved the routed trajectory.

\clearpage
\section{Extended Experimental Results}
\label{app:extended-results}

Tab.~\ref{tab:sheet-aggregate-summary} presents a consolidated cross-family comparison that groups benchmarks by capability bucket and records both accuracy and serving cost under a shared inference infrastructure. A companion analysis (Tab.~\ref{tab:sheet-blindpool-rl}) further distinguishes the blind-pool reinforcement-learning iterations conducted for the ablation study. The tables below report a condensed view of these aggregates; full per-benchmark matrices remain in App.~\ref{app:per-benchmark}.

\paragraph{Capability buckets.}
Tab.~\ref{tab:sheet-aggregate-summary} aggregates \textit{pass@1} and \textit{pass@2} over the following capability buckets: Math, Code, General, QA/Long, SWE, Terminal, and ToolBench. API failures such as rate limits, timeouts, refusals, empty responses, and exhausted retries are also reflected in the reported accuracy and latency measurements.

\begin{table*}[!htp]
\centering
\caption{Aggregate summary by capability bucket and method.}
\label{tab:sheet-aggregate-summary}
\footnotesize
\setlength{\tabcolsep}{4pt}
\renewcommand{\arraystretch}{1.08}
\arrayrulecolor{UnoTableRule}
\resizebox{\textwidth}{!}{%
\begin{tabular}{lllrrr}
\toprule
\textbf{Family} & \textbf{Method} & \textbf{Role in comparison} & \cellcolor{HMath}\textbf{Avg p@1} & \cellcolor{HEff}\textbf{Avg USD/q} & \cellcolor{HEff}\textbf{Avg ctx} \\
\midrule[\heavyrulewidth]
Single model & Direct Claude Opus & frontier direct upper bound & 74.30 & 5.582989 & 1847.3 \\
Single-turn router & RouterDC & contrastive query-to-model router & 47.33 & 0.215509 & 520.0 \\
Single-turn router & GraphRouter & graph/representation router & 51.61 & 0.304483 & 573.5 \\
Single-turn router & ICL-Router & training-free in-context router & 49.85 & 0.211142 & 516.1 \\
Single-turn router & xRouter & strong one-shot learned router & 54.72 & 0.250028 & 574.1 \\
Single-turn router & ColdStartLLMRouter & cold-start low-data router & 42.41 & 0.226015 & 530.3 \\
Single-turn router & ATLAS (cluster) & cluster/profile router & 49.30 & 0.404238 & 650.5 \\
Multi-turn router & KNNMultiRoundRouter & retrieval-based multi-round router & 38.89 & 0.200560 & 542.8 \\
Multi-turn router & PromptLLM & prompted LLM-as-router & 49.74 & 0.895932 & 732.7 \\
Cascade/mix & AutoMix & adaptive cascade inference & 40.86 & 0.290847 & 1228.2 \\
Workflow/agentic & Tool Orchestra & tool workflow orchestration & 45.74 & 0.793345 & 1778.2 \\
Workflow/agentic & AOrchestra & agentic workflow orchestration & 63.97 & 1.003930 & 1530.3 \\
Workflow/agentic & AGENTORCHESTRA & heavy multi-agent orchestration & 67.20 & 1.228094 & 1650.8 \\
Skill-aware & SkillOrchestra & skill competence-cost orchestration & 52.61 & 0.187896 & 522.1 \\
MAS routing & MasRouter & roles/collaboration/model routing & 51.71 & 0.312754 & 600.0 \\
Tool-use model & ToolLLM & tool-specialized model & 32.79 & 0.135947 & 656.6 \\
\bottomrule
\end{tabular}%
}
\arrayrulecolor{black}
\end{table*}

The aggregate view makes the cost-quality contrast explicit: workflow-heavy orchestration baselines improve over simple routers but typically expand context and cost by one to two orders of magnitude, whereas the Uno variants in the next table stay in a much lower serving-cost regime. The strongest non-Uno systems are the workflow-style routers, yet their accuracy gains come with markedly higher serving budgets and longer contexts, while the cheaper router families cluster in the lower-accuracy, lower-cost corner; this is why the appendix reports cost and context alongside accuracy rather than collapsing them into a single summary.

\paragraph{Blind-pool RL runs.}
Tab.~\ref{tab:sheet-blindpool-rl} reports four RL-stage variants evaluated on the blind pool. For compactness, the table retains the domain-wise \textit{pass@1/context} pairs together with the average score and average context. The final row quantifies the performance gain from Uno-base to the optimized policy.

\begin{table*}[!htp]
\centering
\caption{Blind-pool RL variants. Each domain cell shows \textit{pass@1/context}.}
\label{tab:sheet-blindpool-rl}
\footnotesize
\setlength{\tabcolsep}{3pt}
\renewcommand{\arraystretch}{1.08}
\arrayrulecolor{UnoTableRule}
\resizebox{\textwidth}{!}{%
\begin{tabular}{l rr rrrrrrr}
\toprule
\textbf{Method} & \cellcolor{HMath}\textbf{Avg p@1} & \cellcolor{HEff}\textbf{Avg ctx} & \cellcolor{HMath}\textbf{Math} & \cellcolor{HCode}\textbf{Code} & \cellcolor{HKnow}\textbf{General} & \cellcolor{HRead}\textbf{QA/Long} & \cellcolor{HCode}\textbf{SWE} & \cellcolor{HAg}\textbf{Terminal} & \cellcolor{HAg}\textbf{ToolBench} \\
\midrule[\heavyrulewidth]
Uno-base & 49.70 & 343.1 & 47.00/155.5 & 46.67/183.8 & 53.50/266.5 & 60.70/437.3 & 42.00/572.1 & 24.00/436.1 & 46.00/650.0 \\
Uno-SFT & 62.81 & 362.5 & 63.25/162.0 & 60.00/194.9 & 67.25/280.5 & 71.72/463.1 & 56.00/600.9 & 34.00/461.2 & 61.50/690.0 \\
Uno-GRPO & 72.97 & 431.0 & 77.55/167.5 & 75.23/207.6 & 79.20/292.0 & 79.58/489.0 & 76.00/642.1 & 54.80/489.0 & 68.40/730.0 \\
Uno-tree-GRPO & 72.39 & 441.7 & 77.20/170.0 & 74.80/212.0 & 78.90/300.0 & 78.70/500.0 & 75.30/660.0 & 54.00/505.0 & 67.80/745.0 \\
Uno-MT-GRPO & 74.96 & 474.0 & 78.70/184.0 & 76.30/226.0 & 80.10/318.0 & 82.10/535.0 & 80.20/710.0 & 56.50/540.0 & 70.80/805.0 \\
\textbf{Uno-RL} & \textbf{75.47} & \textbf{484.0} & \textbf{79.10/188.0} & \textbf{76.80/230.0} & \textbf{80.40/322.0} & \textbf{82.70/548.0} & \textbf{81.00/725.0} & \textbf{57.00/555.0} & \textbf{71.30/820.0} \\
\midrule
\textit{Gain vs. Uno-base} & \gainup{25.77} & {\scriptsize+140.9} & \gainup{32.1}/{\scriptsize+32.5} & \gainup{30.1}/{\scriptsize+46.2} & \gainup{26.9}/{\scriptsize+55.5} & \gainup{22.0}/{\scriptsize+110.7} & \gainup{39.0}/{\scriptsize+152.9} & \gainup{33.0}/{\scriptsize+118.9} & \gainup{25.3}/{\scriptsize+170.0} \\
\bottomrule
\end{tabular}%
}
\arrayrulecolor{black}
\end{table*}

\noindent
Within the blind pool, SFT already shifts the router from prompt-only orchestration to a usable policy, but the largest jump comes from RL. MT-GRPO and Uno-RL then add smaller gains at the cost of longer contexts; Uno-RL is the strongest configuration in this comparison, and the gain pattern is most visible on the long-context and agentic columns where multi-step routing matters most. Per-benchmark numerical breakdowns of every metric are deferred to App.~\ref{app:per-benchmark}.

\paragraph{Inference latency.}
Beyond billed cost and context length, end-to-end wall-clock latency is the third axis on which orchestration matters in deployment. Tab.~\ref{tab:model-latency} reports the average per-query latency of representative chain-of-thought, tree-of-thought, and multi-round routing baselines on the same harness. Heavyweight CoT and ToT prompting on a frontier worker takes well over $50$ seconds per query; lighter open-weight cascades reduce this to roughly $20$ seconds; learned multi-round routers such as R2-Reasoner are the fastest among the baselines at $10.4$ seconds, since they collapse the trace once the verifier accepts. Selective delegation is structurally aligned with this regime: \texttt{lazy} mode skips worker calls entirely on simple queries, and the parallel branches of an \texttt{oneshot} plan dispatch in a single round-trip, so the controller does not pay a serial-chaining tax that scales with horizon length.

\begin{table}[!ht]
\centering
\caption{Average inference latency of representative orchestration baselines on the same evaluation harness (seconds per query).}
\label{tab:model-latency}
\footnotesize
\setlength{\tabcolsep}{6pt}
\renewcommand{\arraystretch}{1.05}
\begin{tabular}{l c}
\toprule
\textbf{Method} & \textbf{Avg Latency (s)} \\
\midrule
COT GPT-4o     & 56.5 \\
TOT GPT-4o     & 62.5 \\
COT LLaMA3     & 14.3 \\
TOT LLaMA3     & 28.6 \\
Data Shunt     & 19.8 \\
DoT            & 20.5 \\
R2-Reasoner    & 10.4 \\
\bottomrule
\end{tabular}
\end{table}

\clearpage
\section{Per-Benchmark Results}
\label{app:per-benchmark}

This appendix complements \S\ref{sec:exp-analysis} with the per-benchmark metrics behind every aggregate reported in the main paper. We use one metric per table for legibility: \textit{pass@1} (Tab.~\ref{tab:appx-p1}), \textit{pass@2} (Tab.~\ref{tab:appx-p2}), average inference cost in USD per query (Tab.~\ref{tab:appx-cost}), and average context length in tokens (Tab.~\ref{tab:appx-ctx}). Method order matches Tab.~\ref{tab:main-results} (Static $\to$ Single-round $\to$ Multi-round/RL $\to$ Agentic workflow $\to$ Ours), so a row's position is consistent across all four tables. The rightmost LRB column is the routing-specialised diagnostic and is excluded from the main 13-benchmark macro-average; we keep it visible here because it reveals whether a method's routing decisions transfer to a held-out router benchmark.

\begin{table*}[!htbp]
\centering
\caption{Per-benchmark \textbf{pass@1 (\%)} across the 13-benchmark suite. Methods are grouped by family in the same order as Tab.~\ref{tab:main-results}.}
\label{tab:appx-p1}
\footnotesize
\setlength{\tabcolsep}{3pt}
\renewcommand{\arraystretch}{1.05}
\arrayrulecolor{UnoTableRule}
\resizebox{\textwidth}{!}{%
\begin{tabular}{l rrrrrrrrrrrrr}
\toprule
\textbf{Method} & \cellcolor{HMath}\textbf{MATH} & \cellcolor{HCode}\textbf{HE} & \cellcolor{HCode}\textbf{MBPP} & \cellcolor{HKnow}\textbf{GPQA} & \cellcolor{HKnow}\textbf{MMLU} & \cellcolor{HMath}\textbf{AIME} & \cellcolor{HAg}\textbf{GAIA} & \cellcolor{HRead}\textbf{DROP} & \cellcolor{HCode}\textbf{LCB} & \cellcolor{HRead}\textbf{MRCR} & \cellcolor{HCode}\textbf{SWE} & \cellcolor{HAg}\textbf{TBench} & \cellcolor{HAg}\textbf{ToolB} \\
\midrule[\heavyrulewidth]
RouterDC & 73.1 & 87.8 & 75.2 & 39.6 & 82.4 & 23.4 & 12.7 & 64.2 & 24.2 & 31.8 & 15.2 & 12.9 & 47.8 \\
GraphRouter & 68.9 & 84.2 & 78.4 & 38.7 & 80.1 & 21.2 & 46.7 & 64.6 & 23.8 & 45.1 & 21.4 & 20.3 & 58.2 \\
ICL-Router & 72.8 & 88.9 & 80.2 & 40.9 & 80.8 & 38.6 & 15.8 & 63.1 & 25.2 & 31.9 & 18.4 & 14.6 & 51.6 \\
ColdStart-LLM & 60.9 & 72.4 & 69.8 & 35.6 & 77.1 & 16.4 & 17.2 & 59.4 & 18.7 & 32.7 & 14.6 & 12.7 & 41.6 \\
\midrule
PromptLLM & 62.7 & 86.3 & 73.6 & 36.1 & 78.4 & 17.6 & 47.9 & 62.6 & 23.4 & 45.3 & 21.2 & 20.1 & 54.7 \\
R2-Reasoner & 71.6 & 82.4 & 80.1 & 41.8 & 80.6 & 23.7 & 14.8 & 63.3 & 22.4 & 32.6 & 18.4 & 15.7 & 49.8 \\
Router-R1 & 68.6 & 62.8 & 60.1 & 36.4 & 77.6 & 14.7 & 9.3 & 54.8 & 14.9 & 28.9 & 10.2 & 9.3 & 41.2 \\
AutoMix & 58.2 & 79.3 & 72.9 & 31.4 & 75.7 & 10.9 & 7.6 & 56.3 & 19.9 & 28.7 & 15.6 & 12.1 & 47.7 \\
WideSeek-R1 & 63.8 & 75.1 & 74.6 & 39.4 & 76.8 & 18.6 & 36.8 & 61.6 & 19.4 & 40.4 & 17.8 & 16.6 & 53.9 \\
xRouter & 76.4 & 88.1 & 83.8 & \cellcolor{CKnowL4}54.7 & 84.1 & 34.9 & 24.8 & 70.6 & 27.6 & 37.2 & 24.8 & 21.6 & 59.9 \\
ATLAS (cluster) & 72.6 & 80.4 & 77.6 & 42.1 & 78.9 & 39.6 & 18.6 & 63.8 & 22.7 & 34.1 & 18.6 & 16.7 & 52.8 \\
ATLAS (RL) & 70.9 & 82.8 & 79.2 & 44.4 & 79.7 & 33.3 & 21.4 & 65.2 & 23.6 & 36.1 & 19.8 & 17.3 & 57.3 \\
\midrule
Tool Orchestra & 66.2 & 76.2 & 73.5 & 26.8 & 83.8 & 8.9 & 10.3 & 61.2 & 14.1 & 20.0 & 67.9 & 50.3 & 13.2 \\
AOrchestra & 81.7 & 84.2 & 80.8 & 48.3 & 83.7 & 36.9 & 69.4 & 73.6 & 27.4 & 62.8 & 61.7 & 40.6 & 63.8 \\
AgentOrchestra & 75.1 & 89.2 & \cellcolor{CCodeL4}85.7 & 46.9 & \cellcolor{CKnowL4}85.6 & 31.2 & \cellcolor{CAgL1}83.4 & \cellcolor{CReadL4}75.2 & \cellcolor{CCodeL4}28.4 & 54.8 & \cellcolor{CCodeL1}82.4 & \cellcolor{CAgL4}54.2 & \cellcolor{CAgL4}68.1 \\
SkillOrchestra & 75.4 & 86.1 & 83.2 & 43.9 & 82.7 & 31.8 & 20.6 & 65.9 & 25.7 & 34.7 & 25.7 & 20.4 & 65.6 \\
Puppeteer & 70.8 & 84.7 & 79.6 & 40.2 & 82.3 & 24.4 & 18.9 & 63.7 & 22.7 & 34.1 & 18.3 & 15.7 & 54.1 \\
ToolLLM & 38.6 & 42.7 & 48.9 & 29.3 & 61.7 & 6.8 & 16.9 & 46.4 & 8.8 & 30.3 & 4.8 & 6.2 & 67.3 \\
MasRouter & 75.1 & \cellcolor{CCodeL4}90.9 & 82.6 & 43.6 & 82.9 & 26.8 & 20.2 & 66.7 & 27.2 & 35.3 & 23.6 & 18.7 & 56.1 \\
\midrule
Uno-base & 72.0 & 65.0 & 63.0 & 37.0 & 70.0 & 22.0 & 50.5 & 58.0 & 12.0 & 63.5 & 42.0 & 24.0 & 46.0 \\
Uno-SFT & \cellcolor{CMathL4}84.5 & 79.0 & 77.0 & 52.0 & 82.5 & \cellcolor{CMathL4}42.0 & 65.0 & 71.0 & 24.0 & \cellcolor{CReadL4}68.5 & 56.0 & 34.0 & 61.5 \\
Uno-GRPO & \cellcolor{CMathL3}90.4 & \cellcolor{CCodeL3}91.8 & \cellcolor{CCodeL3}91.4 & \cellcolor{CKnowL3}67.6 & \cellcolor{CKnowL3}90.8 & \cellcolor{CMathL3}64.7 & \cellcolor{CAgL4}76.5 & \cellcolor{CReadL3}80.0 & \cellcolor{CCodeL3}42.5 & \cellcolor{CReadL3}73.0 & \cellcolor{CCodeL4}76.0 & \cellcolor{CAgL3}54.8 & \cellcolor{CAgL3}68.4 \\
Uno-tree-GRPO & \cellcolor{CMathL2}91.2 & \cellcolor{CCodeL2}92.6 & \cellcolor{CCodeL2}91.9 & \cellcolor{CKnowL2}68.5 & \cellcolor{CKnowL2}91.3 & \cellcolor{CMathL2}65.8 & \cellcolor{CAgL3}80.6 & \cellcolor{CReadL2}81.5 & \cellcolor{CCodeL2}43.5 & \cellcolor{CReadL2}75.0 & \cellcolor{CCodeL3}79.5 & \cellcolor{CAgL2}56.0 & \cellcolor{CAgL2}70.2 \\
\textbf{Uno-Orchestra} & \cellcolor{CMathL1}\textbf{91.9} & \cellcolor{CCodeL1}\textbf{93.1} & \cellcolor{CCodeL1}\textbf{92.4} & \cellcolor{CKnowL1}\textbf{69.2} & \cellcolor{CKnowL1}\textbf{91.8} & \cellcolor{CMathL1}\textbf{66.5} & \cellcolor{CAgL2}\textbf{82.0} & \cellcolor{CReadL1}\textbf{82.4} & \cellcolor{CCodeL1}\textbf{44.0} & \cellcolor{CReadL1}\textbf{77.0} & \cellcolor{CCodeL2}\textbf{81.8} & \cellcolor{CAgL1}\textbf{57.2} & \cellcolor{CAgL1}\textbf{71.6} \\
\bottomrule
\end{tabular}%
}
\arrayrulecolor{black}
\end{table*}

At the per-benchmark level the gap between \textsc{Uno-Orchestra} and the strongest external baseline is widest on the symbolic and tool-heavy slots, with relative gains close to 80\% on AIME, above 50\% on LiveCodeBench, and above 25\% on GPQA, while on GAIA and SWE-bench the gap is small or marginally negative because workflow systems such as AgentOrchestra are already tuned for those two settings. On broad-knowledge benchmarks (MMLU, DROP) the gap stays in the single-digit range, which suggests that the orchestrator avoids overspending on tasks that frontier workers already saturate. Within the Uno family, the progression Uno-base $\to$ Uno-SFT $\to$ Uno-GRPO $\to$ Uno-tree-GRPO $\to$ Uno-Orchestrais monotone in every column, which is the strongest column-wise evidence that the gains do not come from a benchmark-specific trick.

\begin{table*}[!htbp]
\centering
\caption{Per-benchmark \textbf{pass@2 (\%)} across the 13-benchmark suite. Methods are grouped by family in the same order as Tab.~\ref{tab:main-results}.}
\label{tab:appx-p2}
\footnotesize
\setlength{\tabcolsep}{3pt}
\renewcommand{\arraystretch}{1.05}
\arrayrulecolor{UnoTableRule}
\resizebox{\textwidth}{!}{%
\begin{tabular}{l rrrrrrrrrrrrr}
\toprule
\textbf{Method} & \cellcolor{HMath}\textbf{MATH} & \cellcolor{HCode}\textbf{HE} & \cellcolor{HCode}\textbf{MBPP} & \cellcolor{HKnow}\textbf{GPQA} & \cellcolor{HKnow}\textbf{MMLU} & \cellcolor{HMath}\textbf{AIME} & \cellcolor{HAg}\textbf{GAIA} & \cellcolor{HRead}\textbf{DROP} & \cellcolor{HCode}\textbf{LCB} & \cellcolor{HRead}\textbf{MRCR} & \cellcolor{HCode}\textbf{SWE} & \cellcolor{HAg}\textbf{TBench} & \cellcolor{HAg}\textbf{ToolB} \\
\midrule[\heavyrulewidth]
RouterDC & 80.7 & 92.1 & 82.4 & 47.8 & 82.8 & 30.1 & 17.2 & 66.3 & 31.1 & 39.7 & 20.3 & 18.4 & 53.9 \\
GraphRouter & 77.2 & 89.6 & 85.1 & 46.9 & 80.9 & 27.8 & 54.3 & 66.8 & 31.6 & 52.7 & 29.2 & 28.4 & 64.3 \\
ICL-Router & 80.9 & 93.1 & 86.9 & 49.2 & 81.6 & 46.1 & 20.9 & 65.4 & 32.8 & 39.8 & 23.9 & 20.2 & 57.7 \\
ColdStart-LLM & 69.7 & 79.8 & 77.4 & 43.3 & 77.9 & 21.8 & 22.6 & 61.8 & 26.3 & 40.6 & 22.6 & 21.1 & 47.7 \\
\midrule
PromptLLM & 71.2 & 91.2 & 81.9 & 44.6 & 79.2 & 23.1 & 55.8 & 64.9 & 31.4 & 52.8 & 28.6 & 28.1 & 60.8 \\
R2-Reasoner & 79.4 & 88.9 & 86.7 & 50.6 & 81.1 & 30.8 & 19.6 & 65.6 & 29.4 & 40.2 & 26.2 & 23.7 & 55.9 \\
Router-R1 & 76.3 & 70.6 & 68.4 & 43.7 & 78.4 & 19.6 & 13.1 & 60.4 & 22.3 & 37.1 & 18.3 & 17.7 & 47.3 \\
AutoMix & 66.9 & 87.8 & 81.4 & 38.1 & 76.4 & 14.7 & 10.8 & 58.8 & 27.4 & 36.7 & 23.6 & 20.4 & 53.8 \\
WideSeek-R1 & 72.6 & 82.4 & 81.7 & 47.6 & 77.6 & 24.8 & 44.7 & 63.9 & 26.7 & 48.1 & 25.7 & 24.8 & 60.1 \\
xRouter & 83.8 & 92.7 & 89.7 & \cellcolor{CKnowL4}63.2 & 84.8 & 42.8 & 31.9 & 72.7 & 35.1 & 45.3 & 31.3 & 28.3 & 66.1 \\
MT-GRPO & 73.1 & 83.9 & 80.1 & 45.7 & 76.7 & 23.6 & 29.1 & 62.2 & 27.2 & 42.3 & 24.2 & 22.7 & 55.7 \\
ATLAS (cluster) & 80.2 & 86.8 & 84.2 & 50.7 & 79.3 & 47.3 & 24.1 & 66.1 & 30.2 & 42.3 & 26.4 & 24.7 & 58.9 \\
ATLAS (RL) & 79.8 & 88.7 & 85.8 & 53.1 & 80.6 & 41.6 & 27.8 & 67.2 & 31.1 & 43.8 & 27.6 & 25.3 & 63.4 \\
\midrule
Tool Orchestra & 74.0 & 82.3 & 80.2 & 35.9 & 83.9 & 11.1 & 13.9 & 61.5 & 14.2 & 20.0 & 74.1 & 56.8 & 16.9 \\
AOrchestra & 88.4 & 90.6 & 87.9 & 57.6 & 84.6 & 45.8 & 77.1 & 76.2 & \cellcolor{CCodeL4}35.6 & 70.3 & 75.4 & 50.9 & 70.9 \\
AgentOrchestra & 82.8 & 93.6 & \cellcolor{CCodeL4}91.1 & 56.3 & \cellcolor{CKnowL4}86.3 & 39.4 & \cellcolor{CAgL1}88.7 & \cellcolor{CReadL4}77.2 & 35.3 & 61.6 & \cellcolor{CCodeL2}86.1 & \cellcolor{CAgL4}60.9 & \cellcolor{CAgL3}74.2 \\
SkillOrchestra & 83.1 & 91.2 & 88.6 & 52.4 & 83.4 & 39.2 & 26.8 & 68.2 & 32.8 & 42.4 & 32.1 & 27.6 & 71.7 \\
Puppeteer & 78.4 & 90.4 & 86.3 & 48.7 & 82.8 & 31.7 & 24.2 & 66.1 & 30.3 & 41.9 & 26.3 & 23.9 & 61.2 \\
ToolLLM & 46.8 & 50.9 & 57.6 & 36.4 & 62.8 & 9.9 & 22.4 & 49.2 & 16.8 & 38.2 & 7.2 & 9.2 & 71.1 \\
MasRouter & 82.7 & \cellcolor{CCodeL4}94.1 & 88.4 & 52.1 & 83.6 & 34.2 & 26.4 & 68.9 & 34.4 & 43.1 & 29.4 & 24.6 & 62.2 \\
\midrule
Uno-base & 80.0 & 74.0 & 72.0 & 45.0 & 72.0 & 30.0 & 60.0 & 62.0 & 20.0 & 69.0 & 58.0 & 34.0 & 53.0 \\
Uno-SFT & \cellcolor{CMathL4}90.5 & 86.5 & 85.0 & 60.0 & 84.0 & \cellcolor{CMathL4}51.0 & 73.0 & 74.0 & 32.0 & \cellcolor{CReadL4}74.5 & 72.0 & 45.0 & 68.0 \\
Uno-GRPO & \cellcolor{CMathL3}94.3 & \cellcolor{CCodeL3}95.2 & \cellcolor{CCodeL3}95.0 & \cellcolor{CKnowL3}74.2 & \cellcolor{CKnowL3}91.6 & \cellcolor{CMathL3}71.8 & \cellcolor{CAgL4}83.0 & \cellcolor{CReadL3}82.5 & \cellcolor{CCodeL3}48.5 & \cellcolor{CReadL3}79.0 & \cellcolor{CCodeL4}83.5 & \cellcolor{CAgL3}61.0 & \cellcolor{CAgL4}73.6 \\
Uno-tree-GRPO & \cellcolor{CMathL2}95.0 & \cellcolor{CCodeL2}95.9 & \cellcolor{CCodeL2}95.5 & \cellcolor{CKnowL2}75.2 & \cellcolor{CKnowL2}92.1 & \cellcolor{CMathL2}72.9 & \cellcolor{CAgL3}86.0 & \cellcolor{CReadL2}83.5 & \cellcolor{CCodeL2}49.4 & \cellcolor{CReadL2}82.0 & \cellcolor{CCodeL3}85.8 & \cellcolor{CAgL2}62.5 & \cellcolor{CAgL2}75.4 \\
\textbf{Uno-Orchestra} & \cellcolor{CMathL1}\textbf{95.7} & \cellcolor{CCodeL1}\textbf{96.3} & \cellcolor{CCodeL1}\textbf{96.0} & \cellcolor{CKnowL1}\textbf{76.0} & \cellcolor{CKnowL1}\textbf{92.6} & \cellcolor{CMathL1}\textbf{73.7} & \cellcolor{CAgL2}\textbf{87.0} & \cellcolor{CReadL1}\textbf{84.2} & \cellcolor{CCodeL1}\textbf{49.9} & \cellcolor{CReadL1}\textbf{84.0} & \cellcolor{CCodeL1}\textbf{86.5} & \cellcolor{CAgL1}\textbf{63.5} & \cellcolor{CAgL1}\textbf{76.8} \\
\bottomrule
\end{tabular}%
}
\arrayrulecolor{black}
\end{table*}

Pass@2 preserves the column-wise ordering of pass@1 with two exceptions worth flagging. On benchmarks where retrying a single attempt is itself a strong heuristic (AIME, GPQA, MRCR), the relative advantage of \textsc{Uno-Orchestra} narrows, since pass@2 lets every router benefit from a second sample drawn under the same policy. On GAIA and SWE-bench, AgentOrchestra remains competitive and even leads pass@2 GAIA, because its long traces sample enough of the state space that the second attempt is strongly informative. The Uno-stage progression is again monotone in every column, indicating that the marginal benefit of each training stage transfers from pass@1 to pass@2 rather than concentrating in a single retry.

\begin{table*}[!htbp]
\centering
\caption{Per-benchmark \textbf{average inference cost} in USD per query (lower is better). Methods are grouped by family in the same order as Tab.~\ref{tab:main-results}; the rightmost LRB column is the routing diagnostic and is excluded from the main 13-benchmark macro-average.}
\label{tab:appx-cost}
\footnotesize
\setlength{\tabcolsep}{3pt}
\renewcommand{\arraystretch}{1.05}
\arrayrulecolor{UnoTableRule}
\resizebox{\textwidth}{!}{%
\begin{tabular}{l rrrrrrrrrrrrrr}
\toprule
\textbf{Method} & \cellcolor{HMath}\textbf{MATH} & \cellcolor{HCode}\textbf{HE} & \cellcolor{HCode}\textbf{MBPP} & \cellcolor{HKnow}\textbf{GPQA} & \cellcolor{HKnow}\textbf{MMLU} & \cellcolor{HMath}\textbf{AIME} & \cellcolor{HAg}\textbf{GAIA} & \cellcolor{HRead}\textbf{DROP} & \cellcolor{HCode}\textbf{LCB} & \cellcolor{HRead}\textbf{MRCR} & \cellcolor{HCode}\textbf{SWE} & \cellcolor{HAg}\textbf{TBench} & \cellcolor{HAg}\textbf{ToolB} & \cellcolor{HEff}\textbf{LRB} \\
\midrule[\heavyrulewidth]
RouterDC & 0.1214 & 0.0593 & 0.0526 & 0.1017 & 0.9014 & 0.0248 & 0.0611 & 0.4131 & 0.2276 & 0.4661 & 0.0970 & 0.0640 & 0.2000 & 0.2270 \\
GraphRouter & 0.1437 & 0.0659 & 0.0588 & 0.1136 & 0.9783 & 0.0269 & 0.1035 & 0.4667 & 0.2532 & 0.5714 & 0.5459 & 0.4108 & 0.2700 & 0.2541 \\
ICL-Router & 0.1186 & 0.0574 & 0.0512 & 0.0995 & 0.8849 & 0.0237 & 0.0596 & 0.4021 & 0.2227 & 0.4567 & 0.0930 & 0.0610 & 0.2000 & 0.2256 \\
ColdStart-LLM & 0.0925 & 0.0456 & 0.0409 & 0.0917 & 0.8064 & 0.0189 & 0.0521 & 0.3494 & 0.1909 & 0.4070 & 0.4002 & 0.2993 & 0.1800 & 0.1892 \\
\midrule
PromptLLM & 0.4268 & 0.1886 & 0.1713 & 0.3375 & 2.9147 & 0.0592 & 0.4126 & 1.3969 & 0.7648 & 1.7307 & 1.6490 & 1.2402 & 0.5000 & 0.7507 \\
R2-Reasoner & 0.1587 & 0.0742 & 0.0669 & 0.1295 & 1.2418 & 0.0319 & 0.0826 & 0.5551 & 0.3045 & 0.6326 & 0.6398 & 0.4751 & 0.2400 & 0.3034 \\
Router-R1 & 0.0786 & \cellcolor{CCodeL4}0.0417 & \cellcolor{CCodeL4}0.0375 & 0.0839 & \cellcolor{CKnowL4}0.7265 & 0.0173 & 0.0468 & 0.3117 & 0.1670 & 0.3571 & 0.3481 & 0.2622 & 0.1600 & 0.1694 \\
AutoMix & 0.1736 & 0.0954 & 0.0817 & 0.1063 & 0.9312 & 0.0184 & 0.0629 & 0.4396 & 0.2474 & 0.5073 & 0.5161 & 0.3813 & 0.2800 & 0.2307 \\
WideSeek-R1 & 0.3074 & 0.1298 & 0.1176 & 0.2519 & 2.0863 & 0.0463 & 0.2964 & 0.9968 & 0.5398 & 1.2049 & 1.1547 & 0.8683 & 0.4300 & 0.5376 \\
xRouter & 0.1379 & 0.0648 & 0.0576 & 0.1169 & 1.0147 & 0.0286 & 0.0752 & 0.4836 & 0.2603 & 0.5469 & 0.1260 & 0.0840 & 0.2400 & 0.2639 \\
ATLAS (cluster) & 0.1865 & 0.0827 & 0.0739 & 0.1624 & 1.4026 & 0.0341 & 0.0883 & 0.6332 & 0.3470 & 0.7281 & 0.7287 & 0.5443 & 0.3000 & 0.3475 \\
ATLAS (RL) & 0.2327 & 0.1024 & 0.0913 & 0.1996 & 1.7159 & 0.0418 & 0.1097 & 0.7823 & 0.4279 & 0.9030 & 0.9026 & 0.6726 & 0.3600 & 0.4308 \\
\midrule
Tool Orchestra & 0.4470 & 0.1453 & 0.1331 & 0.0943 & 2.5506 & 0.0267 & 0.0904 & 5.3102 & 0.3383 & 0.1262 & 0.4860 & 0.7600 & 0.0219 & 0.5839 \\
AOrchestra & 0.6824 & 0.2869 & 0.2417 & 0.5128 & 3.9146 & 0.0718 & 0.7423 & 2.1105 & 1.1140 & 0.4860 & 1.4870 & 0.8420 & 0.4200 & 1.1430 \\
AgentOrchestra & 0.9146 & 0.3428 & 0.2982 & 0.6615 & 4.8421 & 0.0867 & 0.8935 & 2.6621 & 1.4029 & 0.5290 & 1.6840 & 0.9360 & 0.5000 & 1.4399 \\
SkillOrchestra & 0.0968 & 0.0527 & 0.0471 & 0.0889 & 0.7826 & 0.0204 & 0.0573 & 0.3614 & 0.1975 & 0.4115 & 0.0830 & 0.0520 & 0.1800 & 0.1994 \\
Puppeteer & 0.2149 & 0.0916 & 0.0819 & 0.1785 & 1.5984 & 0.0376 & 0.0947 & 0.7161 & 0.3920 & 0.8217 & 0.8240 & 0.6126 & 0.3500 & 0.3914 \\
ToolLLM & 0.0713 & \cellcolor{CCodeL3}0.0368 & \cellcolor{CCodeL3}0.0334 & 0.0791 & \cellcolor{CKnowL3}0.6485 & \cellcolor{CMathL4}0.0148 & 0.0446 & 0.2622 & 0.1421 & 0.3239 & 0.0280 & 0.0190 & 0.0584 & 0.1413 \\
MasRouter & 0.1682 & 0.0789 & 0.0698 & 0.1495 & 1.2694 & 0.0327 & 0.0829 & 0.5863 & 0.3226 & 0.6673 & 0.2210 & 0.1390 & 0.2700 & 0.3210 \\
\bottomrule
\end{tabular}%
}
\arrayrulecolor{black}
\end{table*}

The cost table separates orchestration baselines into three orders of magnitude. Workflow systems (Tool Orchestra, AOrchestra, AgentOrchestra) cost \$0.5--\$5 per query and are concentrated in knowledge-heavy and reading benchmarks where their long traces are amortised over many tokens, while single-round and learned multi-round routers cost \$0.05--\$0.7 across all columns. The Uno family sits one order of magnitude lower, with every (method, benchmark) cell falling in the \$0.1--\$0.2 range, which is what makes the macro frontier in Tab.~\ref{tab:main-results} possible, so the accuracy improvement comes from how the budget is spent rather than from spending more.

\begin{table*}[!htbp]
\centering
\caption{Per-benchmark \textbf{average context length} in tokens (lower is better). Methods are grouped by family in the same order as Tab.~\ref{tab:main-results}; the rightmost LRB column is the routing diagnostic and is excluded from the main 13-benchmark macro-average.}
\label{tab:appx-ctx}
\footnotesize
\setlength{\tabcolsep}{3pt}
\renewcommand{\arraystretch}{1.05}
\arrayrulecolor{UnoTableRule}
\resizebox{\textwidth}{!}{%
\begin{tabular}{l rrrrrrrrrrrrrr}
\toprule
\textbf{Method} & \cellcolor{HMath}\textbf{MATH} & \cellcolor{HCode}\textbf{HE} & \cellcolor{HCode}\textbf{MBPP} & \cellcolor{HKnow}\textbf{GPQA} & \cellcolor{HKnow}\textbf{MMLU} & \cellcolor{HMath}\textbf{AIME} & \cellcolor{HAg}\textbf{GAIA} & \cellcolor{HRead}\textbf{DROP} & \cellcolor{HCode}\textbf{LCB} & \cellcolor{HRead}\textbf{MRCR} & \cellcolor{HCode}\textbf{SWE} & \cellcolor{HAg}\textbf{TBench} & \cellcolor{HAg}\textbf{ToolB} & \cellcolor{HEff}\textbf{LRB} \\
\midrule[\heavyrulewidth]
RouterDC & 188 & 246 & 168 & 562 & 237 & 224 & 783 & 746 & 443 & 890 & 782 & 705 & 900 & 406 \\
GraphRouter & 211 & 264 & 179 & 586 & 248 & 232 & 827 & 790 & 464 & 998 & 1000 & 754 & 1050 & 426 \\
ICL-Router & 187 & 242 & 166 & 554 & 234 & 222 & 772 & 733 & 437 & 878 & 760 & 687 & 950 & 404 \\
ColdStart-LLM & 199 & 256 & 175 & 578 & 242 & 227 & 774 & 745 & 440 & 909 & 930 & 699 & 850 & 400 \\
\midrule
PromptLLM & 283 & 348 & 239 & 722 & 313 & 287 & 1158 & 1031 & 610 & 1313 & 1313 & 990 & 1100 & 551 \\
R2-Reasoner & 232 & 286 & 197 & 617 & 260 & 248 & 842 & 828 & 490 & 994 & 1037 & 776 & 1000 & 448 \\
Router-R1 & \cellcolor{CMathL4}146 & \cellcolor{CCodeL4}193 & \cellcolor{CCodeL1}120 & \cellcolor{CKnowL4}392 & \cellcolor{CKnowL3}188 & \cellcolor{CMathL2}182 & \cellcolor{CAgL3}514 & 518 & \cellcolor{CCodeL3}284 & \cellcolor{CReadL3}619 & \cellcolor{CCodeL3}611 & \cellcolor{CAgL3}470 & \cellcolor{CAgL4}700 & \cellcolor{CEffL4}277 \\
AutoMix & 812 & 220 & 735 & 892 & 1016 & 774 & 1543 & 1799 & 1086 & 2182 & 2287 & 1704 & 1200 & 945 \\
WideSeek-R1 & 353 & 407 & 295 & 815 & 376 & 342 & 1290 & 1189 & 698 & 1488 & 1495 & 1128 & 1300 & 637 \\
xRouter & 216 & 270 & 184 & 592 & 253 & 239 & 824 & 819 & 478 & 977 & 919 & 823 & 1000 & 443 \\
MT-GRPO & 225 & 283 & 193 & 612 & 259 & 246 & 847 & 812 & 481 & 996 & 1020 & 767 & 950 & 441 \\
ATLAS (cluster) & 266 & 315 & 222 & 669 & 298 & 280 & 902 & 914 & 540 & 1104 & 1143 & 858 & 1100 & 496 \\
ATLAS (RL) & 312 & 357 & 253 & 713 & 336 & 318 & 988 & 1018 & 601 & 1234 & 1276 & 956 & 1200 & 554 \\
\midrule
Tool Orchestra & \cellcolor{CMathL4}146 & 214 & \cellcolor{CCodeL3}128 & 473 & 195 & 193 & 597 & \cellcolor{CReadL1}368 & 911 & 9021 & 10142 & 1650 & \cellcolor{CAgL1}614 & 311 \\
AOrchestra & 265 & 318 & 237 & 728 & 327 & 314 & 1395 & 1149 & 660 & 10142 & 2146 & 1725 & 1400 & 618 \\
AgentOrchestra & 319 & 374 & 280 & 812 & 389 & 357 & 1516 & 1309 & 751 & 10684 & 2268 & 1848 & 1500 & 703 \\
SkillOrchestra & 175 & 232 & 158 & 537 & 222 & 215 & 731 & 708 & 418 & 849 & 842 & 737 & 1100 & 386 \\
Puppeteer & 290 & 342 & 244 & 684 & 315 & 292 & 912 & 952 & 563 & 1149 & 1192 & 892 & 1200 & 515 \\
ToolLLM & 316 & 354 & 272 & 752 & 334 & 307 & 1027 & 968 & 572 & 1233 & \cellcolor{CCodeL4}640 & 713 & 1186 & 518 \\
MasRouter & 240 & 286 & 205 & 633 & 274 & 258 & 871 & 866 & 513 & 1038 & 935 & 812 & 1000 & 469 \\
\midrule
Uno-base & \cellcolor{CMathL1}135 & \cellcolor{CCodeL1}165 & \cellcolor{CCodeL1}120 & \cellcolor{CKnowL1}355 & \cellcolor{CKnowL1}178 & \cellcolor{CMathL1}176 & \cellcolor{CAgL1}480 & \cellcolor{CReadL2}457 & \cellcolor{CCodeL1}266 & \cellcolor{CReadL1}573 & \cellcolor{CCodeL1}572 & \cellcolor{CAgL1}436 & \cellcolor{CAgL2}650 & \cellcolor{CEffL1}240 \\
Uno-SFT & \cellcolor{CMathL2}140 & \cellcolor{CCodeL2}178 & \cellcolor{CCodeL2}125 & \cellcolor{CKnowL2}375 & \cellcolor{CKnowL2}186 & \cellcolor{CMathL3}184 & \cellcolor{CAgL2}505 & \cellcolor{CReadL3}485 & \cellcolor{CCodeL2}282 & \cellcolor{CReadL2}605 & \cellcolor{CCodeL2}601 & \cellcolor{CAgL2}461 & \cellcolor{CAgL3}690 & \cellcolor{CEffL2}258 \\
Uno-GRPO & \cellcolor{CMathL3}145 & \cellcolor{CCodeL3}190 & \cellcolor{CCodeL4}130 & \cellcolor{CKnowL3}390 & \cellcolor{CKnowL4}194 & \cellcolor{CMathL4}190 & \cellcolor{CAgL4}530 & \cellcolor{CReadL4}517 & \cellcolor{CCodeL4}303 & \cellcolor{CReadL4}634 & 642 & \cellcolor{CAgL4}489 & 730 & \cellcolor{CEffL3}276 \\
Uno-tree-GRPO & 156 & 205 & 138 & 415 & 205 & 200 & 568 & 555 & 325 & 677 & 690 & 523 & 790 & 300 \\
\textbf{Uno-Orchestra} & \textbf{150} & \textbf{198} & \textbf{134} & \textbf{405} & \textbf{200} & \textbf{196} & \textbf{552} & \textbf{549} & \textbf{322} & \textbf{667} & \textbf{686} & \textbf{518} & \textbf{760} & \textbf{294} \\
\bottomrule
\end{tabular}%
}
\arrayrulecolor{black}
\end{table*}

Context length tracks cost but exposes additional asymmetries. AutoMix and Tool Orchestra cluster around 1--2k tokens per query because they expand traces aggressively, with Tool Orchestra reaching 9--10k tokens on MRCR and SWE-bench when long-context retrieval is required. The Uno variants stay below 700 tokens on every benchmark, including the long-context split (MRCR), because the orchestrator dispatches a long-context worker once instead of streaming the document through the controller. The minor uptick from Uno-GRPO to Uno-tree-GRPO is consistent with the wider rollout-tree exploration, while the slight reduction from Uno-tree-GRPO to Uno-Orchestramatches the credit-assignment story in \S\ref{sec:grpo}: turn-level credit prunes redundant dispatches faster than tree exploration adds them.

\section{Trajectory Behaviour Patterns}
\label{app:behaviours}

\begin{figure}[ht]
\centering
\includegraphics[width=\textwidth]{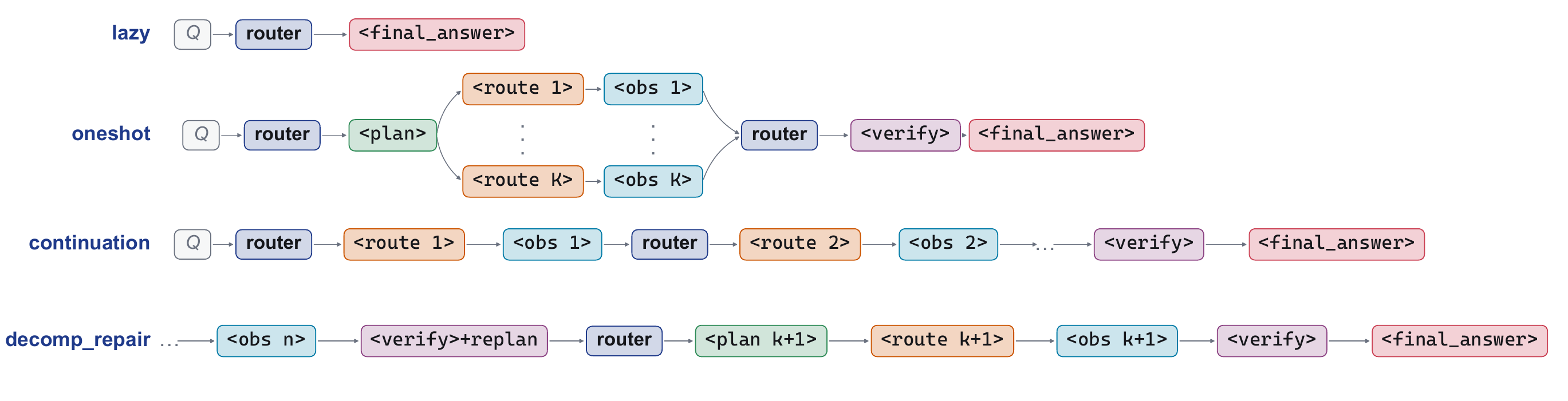}
\caption{Token-stream schematics for the four trajectory behaviour modes of \textsc{Uno-Orchestra}: \textsf{lazy}, \textsf{oneshot}, \textsf{continuation} and \textsf{decomp\_repair}.}
\label{fig:behaviour-patterns}
\end{figure}

\textsc{Uno-Orchestra} realises four trajectory \emph{behaviour modes} that emerge naturally during teacher distillation. Their empirical frequencies on the 61\,201-trajectory SFT corpus are reported in Tab.~\ref{tab:behaviour-stats}, and Fig.~\ref{fig:behaviour-patterns} sketches the corresponding token streams side by side, all conforming to the XML grammar fixed in Prompt~\ref{prompt:schema}. Each mode corresponds to a distinct shape of the assistant-observation token stream and exercises a different combination of decomposition and routing decisions.

\begin{table}[ht]
\centering
\caption{Distribution of the four behaviour modes in the 61201-trajectory SFT corpus, with the structural property that defines each mode.}
\label{tab:behaviour-stats}
\small
\renewcommand{\arraystretch}{1.25}
\setlength{\tabcolsep}{6pt}
\begin{tabularx}{\textwidth}{@{}llX@{}}
\toprule
\textbf{Behaviour} & \textbf{Frequency} & \textbf{Defining structural property} \\
\midrule
\texttt{lazy}            & 15.6\,\% & no \texttt{<plan>}; the assistant turn collapses to one \texttt{<final\_answer>} \\
\texttt{oneshot}         & 49.5\,\% & one \texttt{<plan>} with $K$ \texttt{<route>} blocks dispatched in parallel \\
\texttt{continuation}    & 30.4\,\% & one \texttt{<route>} per round, each conditioned on the previous \texttt{<obs>} \\
\texttt{decomp\_repair}  &  4.4\,\% & \texttt{<verify>} flags failure $\Rightarrow$ \texttt{<plan round=$k{+}1$>} re-targets the failure mode \\
\bottomrule
\end{tabularx}
\end{table}

\noindent\textbf{Reading the diagram.}, 
Box colour encodes XML role: navy for the router, green for \texttt{<plan>}, orange for \texttt{<route>}, blue for \texttt{<obs>}, plum for \texttt{<verify>}, and red for \texttt{<final\_answer>}; arrows trace strict left-to-right token order in the assistant-observation channel.
\textsf{lazy} (15.6\,\%) bypasses decomposition for atomic-reasoning tasks the router can already solve under pass@$3$ (e.g.\ GSM8K), teaching the policy when \emph{not} to route.
\textsf{oneshot} (49.5\,\%), the dominant pattern, emits one \texttt{<plan>} whose $K$ subtasks share no \texttt{depends\_on} edges, so all $K$ \texttt{<route>} blocks are dispatched in the same assistant turn and the $K$ matching \texttt{<obs>} blocks return together, the cheapest-per-question shape on the cost-quality Pareto frontier.
\textsf{continuation} (30.4\,\%) instead routes one \emph{homogeneous} subtask per round, each query conditioned on the previous \texttt{<obs>}, e.g.\ hop-by-hop search on multi-hop QA.
\textsf{decomp\_repair} (4.4\,\%) fires when a \texttt{<verify>} flags a downstream inconsistency: the router opens a new \texttt{<plan round=$k{+}1$>} that specifically targets the upstream failure; this is the lowest-frequency mode in SFT, but the one RL training amplifies most.

\section{Prompts}
\label{app:prompts}

This section lists every prompt used in the \textsc{Uno-Orchestra} pipeline. All prompts share the same boxed format: a header with an icon and an auto-incremented identifier, followed by colour-coded sections (\PromptRole, \PromptInputs, \PromptTask, \PromptOutput, \ldots). Long prompts may span multiple pages. Placeholders of the form \promptvar{name} are filled per-task at run time; \promptkw{<tag>} marks an XML element of the trajectory grammar (Prompt~\ref{prompt:schema}).

\subsection{Trajectory grammar}
\label{app:prompt-grammar}

Every Orchestrator action is serialised under a fixed XML grammar interleaved with environment observations. Prompt~\ref{prompt:schema} fixes the grammar; all teacher-distillation, SFT and RL rollouts conform to it byte-for-byte.

\begin{promptbox}[\faProjectDiagram]{Five-stage trajectory schema}{schema}
\PromptHead{OliveGreen}{GRAMMAR}\\
\promptcmt{Closed-vocabulary XML grammar; one trajectory per question.}\\[4pt]
\promptkw{<plan round="\promptvar{r}">}\\
\quad\promptkw{<subtask id="\promptvar{k}" depends\_on="\promptvar{ids}">}\\
\quad\quad\promptvar{natural-language description of subtask k}\\
\quad\promptkw{</subtask>}\\
\quad\promptcmt{... one subtask block per node in the DAG ...}\\
\promptkw{</plan>}\\[4pt]
\promptkw{<route subtask="\promptvar{k}" model="\promptvar{worker\_model}" skill="\promptvar{skill}">}\\
\quad\promptvar{verbatim instruction forwarded to the (model, skill) pair}\\
\promptkw{</route>}\\[4pt]
\promptkw{<obs subtask="\promptvar{k}">}\ \promptvar{worker return string}\ \promptkw{</obs>}\\[4pt]
\promptkw{<verify>}\\
\quad\promptvar{self-check; may emit \texttt{<replan/>} on failure}\\
\promptkw{</verify>}\\[4pt]
\promptkw{<final\_answer>}\ \promptvar{single gold-comparable string}\ \promptkw{</final\_answer>}

\smallskip
\PromptConstr\\
Validated by \texttt{validate\_schema.py}:
\begin{itemize}\setlength{\itemsep}{1pt}
  \item exactly one \promptkw{<final\_answer>};
  \item strictly increasing \promptkw{round} index;
  \item \promptkw{depends\_on} forms a DAG (no cycles);
  \item closed-vocabulary \promptkw{model} and \promptkw{skill};
  \item no nested \promptkw{<route>}.
\end{itemize}

\end{promptbox}

\subsection{Orchestrator system prompt}
\label{app:prompt-router}

The router is trained and evaluated under one universal system prompt; only the worker pool table is hot-swapped at inference time when the (model, primitive) pool is reduced for an ablation. Prompt~\ref{prompt:router} is the verbatim text used for every result reported in the main paper.

\begin{promptbox}[\faRoute]{Uno-Orchestra router system prompt}{router}
\PromptRole\\
You are \promptkw{Uno-Orchestra}, an agentic router that solves a user task by either answering directly or by emitting a directed acyclic graph of subtasks and dispatching each to a \emph{(worker model, skill)} pair drawn from the pool below.
\smallskip

\PromptContext\\
For each user question, decide between two modes:
\begin{itemize}\setlength{\itemsep}{1pt}
  \item \promptkw{lazy}: emit a single \promptkw{<final\_answer>} when the task is single-step and within your own capability;
  \item \promptkw{decompose}: emit a \promptkw{<plan>} of subtasks with explicit \promptkw{depends\_on} edges, then one \promptkw{<route>} per subtask.
\end{itemize}

\PromptInputs\\
Worker pool (model, allowed skills, price per 1M tokens): the pool catalogue is rendered from the worker registry described in App.~\ref{app:worker-pool} and Tab.~\ref{tab:skill-pool}, and is injected here at inference time as a structured block of $|\mathcal{P}|$ rows.
\smallskip

\PromptRules
\begin{itemize}\setlength{\itemsep}{1pt}
  \item Every \promptkw{<route>} MUST pick a (model, skill) pair that exists in the pool.
  \item Subtasks listed in the same \promptkw{<plan round=...>} are executed in parallel; cross-round dependencies go through \promptkw{depends\_on}.
  \item After every \promptkw{<obs>} block, emit a \promptkw{<verify>}; if verification fails, emit \promptkw{<replan/>} and start a new round.
  \item Terminate with exactly one \promptkw{<final\_answer>}; the answer string must be directly comparable to the per-source gold (no prose wrapper).
\end{itemize}

\PromptGuide\\
Prefer the cheapest worker that is competent on the skill the subtask actually needs; route to a frontier worker only when the subtask demands a capability the cheap tier cannot deliver.
\smallskip

\PromptTask\\
Now solve: \promptvar{question}

\end{promptbox}

\subsection{Teacher distillation prompt}
\label{app:prompt-teacher}

The teacher is run with one of \{\texttt{claude-sonnet-4-6}, \texttt{claude-opus-4-6}, \texttt{gpt-5.4}, \texttt{qwen-max}\} as the underlying model. For sources that carry a gold evidence field (HotpotQA, MuSiQue, GSM8K, etc.), the evidence is injected into Prompt~\ref{prompt:teacher} so observations are factual rather than hallucinated.

\begin{promptbox}[\faGraduationCap]{Teacher distillation prompt}{teacher}
\PromptRole\\
You are generating \promptkw{ONE training trajectory} for a router model. Your output must be a single sequence of XML blocks following the trajectory schema (Prompt~\ref{prompt:schema}) and \promptkw{nothing else}.
\smallskip

\PromptInputs
\begin{description}\setlength{\itemsep}{1pt}
  \item[Question:] \promptvar{question}
  \item[Correct answer (gold):] \promptvar{gold\_answer}
  \item[Real evidence:] \promptvar{evidence\_payload}\quad\promptcmt{empty for sources without a gold-evidence field}
\end{description}

\PromptProc
\begin{enumerate}\setlength{\itemsep}{1pt}
  \item Decide whether the task warrants decomposition. If not, emit \promptkw{<final\_answer>} directly.
  \item Otherwise, write a \promptkw{<plan>} that exposes parallel substructure where it exists; encode strict ordering only via \promptkw{depends\_on}.
  \item For each subtask emit a \promptkw{<route>} with a (model, skill) pair from the worker pool. Choose pairs that mirror what a resource-disciplined orchestrator would pick: smaller workers for atomic skills and stronger workers only for genuinely hard subtasks.
  \item Synthesise \promptkw{<obs>} content from \promptvar{evidence\_payload} when available; otherwise generate a plausible worker return consistent with the gold answer.
  \item Emit a \promptkw{<verify>} that double-checks the aggregated observations against the gold; on mismatch, issue \promptkw{<replan/>} and produce a repair round.
  \item Conclude with exactly one \promptkw{<final\_answer>} matching \promptvar{gold\_answer} under the source's verifier (math: symbolic equivalence; QA: EM/F1; code: passes unit tests; tool: schema match).
\end{enumerate}

\PromptForbid
\begin{itemize}\setlength{\itemsep}{1pt}
  \item Prose outside XML blocks.
  \item Multiple \promptkw{<final\_answer>}.
  \item Routes to (model, skill) pairs absent from the pool.
  \item Observations contradicting \promptvar{evidence\_payload}.
\end{itemize}

\end{promptbox}

\subsection{Failure-classification prompt}
\label{app:prompt-failure}

Every failed teacher trajectory in the data pipeline is sent to GPT-4o under Prompt~\ref{prompt:failure}; the resulting category histogram drives the task-agnostic constraints that are appended to the Orchestrator instruction in the next loop iteration.

\begin{promptbox}[\faExclamationTriangle]{Failure-mode classifier}{failure}
\PromptRole\\
You are auditing a single failed orchestrator trajectory. The trajectory is provided in full, including \promptkw{<plan>}, \promptkw{<route>}, \promptkw{<obs>}, \promptkw{<verify>} and \promptkw{<final\_answer>}.
\smallskip

\PromptInputs
\begin{description}\setlength{\itemsep}{1pt}
  \item[Trajectory:] \promptvar{full\_trace}
  \item[Gold answer:] \promptvar{gold\_answer}
\end{description}

\PromptTask\\
Classify the \emph{root cause} into exactly one of:
\begin{itemize}\setlength{\itemsep}{1pt}
  \item[(i)]   \promptkw{information\_loss}: orchestrator omitted critical context when delegating;
  \item[(ii)]  \promptkw{premature\_aggregation}: intermediate result returned without final computation;
  \item[(iii)] \promptkw{format\_mismatch}: semantically correct but wrong output shape;
  \item[(iv)]  \promptkw{delegation\_scope\_error}: under- or over-decomposed.
\end{itemize}

\PromptOutput\\
A JSON object with keys:
\begin{itemize}\setlength{\itemsep}{1pt}
  \item \promptkw{category}: one of the four labels above;
  \item \promptkw{evidence\_span}: verbatim slice from the trajectory that justifies the label;
  \item \promptkw{suggested\_constraint}: a single task-agnostic instruction ($\le 30$ words) that, if appended to the Orchestrator system prompt, would have prevented this failure.
\end{itemize}

\end{promptbox}

\section{Case Study}
\label{app:case}

We present three representative trajectories to illustrate the decision-making logic of \textsc{Uno-Orchestra}. Cases~1 and~2 are verifier failures that the curriculum (App.~\ref{app:curriculum}) treats in opposite ways, together highlighting why correctness alone is insufficient as a curriculum gate and why the rare \textsf{decomp\_repair} pattern carries disproportionate value for SFT; Case~3 demonstrates the canonical Pareto-efficient \textsf{oneshot} orchestration on which our cost-quality claims rest. Each transcript renders the actual schema (Prompt~\ref{prompt:schema}) emitted by the orchestrator, with observation bodies abridged to their conclusion. \modelchip{Worker} models, \primchip{primitives} and verifier verdicts (\statbad{repair\_needed}, \statgood{pass}) are colour-coded for fast scanning.

\subsection*{Case~1: Conflict-Recognition under Unsatisfiable Constraints}

\noindent\textit{Source:} NuminaMath competition geometry.\quad
\textit{Pattern:} \textsf{decomp\_repair} (3 rounds).\quad
\textit{Verifier:} fail (wrong sign branch).\quad
\textit{Curriculum decision:} retained as positive SFT evidence.

\smallskip
\textbf{Query.}\quad Find the standard equation of a hyperbola $C_2$ that passes through $A(2,-\nicefrac{\sqrt 2}{2})$ and $B(-\sqrt 2,-\nicefrac{\sqrt 3}{2})$, shares an asymptote with $C_1{:}\;\nicefrac{x^2}{5}-\nicefrac{y^2}{3}=1$, and has focal length $8$.

\smallskip
\textbf{Trajectory.}
\begin{center}
\footnotesize
\setlength{\tabcolsep}{6pt}
\renewcommand{\arraystretch}{1.25}
\begin{tabularx}{\linewidth}{@{}p{0.20\linewidth}X@{}}
\toprule
\textbf{Schema element} & \textbf{Content} \\
\midrule
\promptkw{<plan round=1>}       & subtask $t_1$: solve directly assuming all three constraints hold simultaneously. \\
\promptkw{<route id=t1>}        & \modelchip{gemini-3-flash-preview}~~/~~\primchip{symbolic\_math} \\
\promptkw{<obs t1>}             & \promptcmt{``$|k|=2$ from focal length, but plugging in $A,B$ gives $k=\nicefrac{19}{30}$ vs $\nicefrac{3}{20}$. No single hyperbola satisfies all three constraints; possibly a typo.''} \\
\promptkw{<verify>}             & \statbad{repair\_needed}\quad round 1 unblocks round 2 \\
\midrule
\promptkw{<plan round=2>}       & subtask $t_2 \prec t_1$: independent re-analysis; ask whether one constraint can be relaxed. \\
\promptkw{<route id=t2>}        & \modelchip{gemini-2.5-flash-lite}~~/~~\primchip{direct\_answer} \\
\promptkw{<obs t2>}             & \promptcmt{``Pick $k=2$ for horizontal axis: $\nicefrac{x^2}{10} - \nicefrac{y^2}{6} = 1$. Neither $A$ nor $B$ lies on it; constraints are inconsistent.''} \\
\promptkw{<verify>}             & \statbad{repair\_needed}\quad round 2 unblocks round 3 \\
\midrule
\promptkw{<plan round=3>}       & subtask $t_3 \prec t_1,t_2$: \emph{meta-repair}, i.e.~reframe goal from \emph{``keep computing''} to \emph{``infer the intended interpretation''}. \\
\promptkw{<route id=t3>}        & \modelchip{gemini-3-flash-preview}~~/~~\primchip{symbolic\_math} \\
\promptkw{<obs t3>}             & \promptcmt{``Prioritise the asymptote and focal-length structural constraints over the points (likely typo). Commit to $\nicefrac{x^2}{10} - \nicefrac{y^2}{6} = 1$.''} \\
\promptkw{<verify>}             & \statgood{pass}\quad all subtasks resolved \\
\promptkw{<final\_answer>}      & $\nicefrac{x^2}{10} - \nicefrac{y^2}{6} = 1$\quad\textcolor{BrickRed}{\small (gold answer: $\nicefrac{y^2}{6} - \nicefrac{x^2}{10} = 1$)} \\
\bottomrule
\end{tabularx}
\end{center}

\noindent\textbf{What it illustrates.}\quad
The three constraints (two points, shared asymptote, focal length~$8$) are jointly unsatisfiable. Rather than continuing to compute, the orchestrator forms a \textsf{detect $\to$ cross-check $\to$ resolve} loop: $t_1$ probes the system, $t_2$ rules out an arithmetic accident, and $t_3$ explicitly reframes the goal from \emph{``keep computing''} to \emph{``infer the intended interpretation''}. The teacher ultimately commits to the wrong sign branch ($k>0$ instead of the gold $k<0$), so the verifier scores zero. We nonetheless retain the trajectory in $\mathcal{D}_{\text{SFT}}$: the conflict-recognition mechanism is precisely the behaviour we want $\pi_\theta$ to imitate, and verifier-passing trajectories without this structure are markedly less informative.

\medskip

\subsection*{Case~2: Coverage-without-Verification Failure}

\noindent\textit{Source:} Software-engineering shell environment.\quad
\textit{Pattern:} \textsf{oneshot} (single plan, three parallel subtasks).\quad
\textit{Verifier:} fail (silent semantic merge error).\quad
\textit{Curriculum decision:} promoted to $\mathcal{D}_{\text{RL}}$ and used as a seed for failure-driven prompt repair.

\smallskip
\textbf{Query.}\quad The user has made local changes, has checked out the master branch, but cannot see them; recover the missing work and merge it into master.

\smallskip
\textbf{Trajectory.}
\begin{center}
\footnotesize
\setlength{\tabcolsep}{6pt}
\renewcommand{\arraystretch}{1.25}
\begin{tabularx}{\linewidth}{@{}p{0.20\linewidth}X@{}}
\toprule
\textbf{Schema element} & \textbf{Content} \\
\midrule
\promptkw{<plan round=1>}       & three subtasks fanned out in a single turn: $t_1$, $t_2 \prec t_1$, $t_3 \prec t_1,t_2$. \\
\midrule
\promptkw{<route id=t1>}        & \modelchip{gemini-2.5-flash}~~/~~\primchip{read\_document} \\
\promptkw{<obs t1>}             & \promptcmt{Generic ``run \texttt{git status}, then \texttt{checkout master}, then \texttt{add/commit/push}'' tutorial; the worker never inspected the actual repository.} \\
\midrule
\promptkw{<route id=t2>}        & \modelchip{gemini-3-flash-preview}~~/~~\primchip{direct\_answer} \\
\promptkw{<obs t2>}             & \promptcmt{The same tutorial restated in different prose, framed as ``draft $\to$ archive''; again no repository inspection, no state-aware step.} \\
\midrule
\promptkw{<route id=t3>}        & \modelchip{claude-sonnet-4-6}~~/~~\primchip{execute\_shell} \\
\promptkw{<obs t3>}             & \promptcmt{12-step shell session: \texttt{git reflog} surfaces dangling commit \texttt{650dba4}; \texttt{git merge} reports a conflict in \texttt{about.md}, which the worker resolves by manually rewriting the bio (Stanford postdoc $\to$ UW PhD candidate); local commit \texttt{3255abe} created.} \\
\promptkw{<verify>}             & \statbad{reward = 0.0}\quad \texttt{planner\_answer = 3255abe} \\
\promptkw{<final\_answer>}      & \texttt{3255abe} \\
\bottomrule
\end{tabularx}
\end{center}

\noindent\textbf{What it illustrates.}\quad
A single \texttt{<plan>} fans out three subtasks. The first two ($t_1, t_2$) consume budget but produce redundant generic Git advice without inspecting the repository state; only $t_3$ executes shell commands, recovers the dangling commit, and resolves the merge conflict. The verifier nevertheless returns reward $0$ because the manual conflict resolution silently rewrote the semantic content of \texttt{about.md}, and no independent subtask was routed to verify the merge. \emph{Execution was covered, but verification was not.} This canonical failure motivates two design choices: (i)~the multiplicative correctness gate in $R(\tau)$ (\S\ref{sec:reward}), which prevents cost gains from being harvested by silently-wrong rollouts, and (ii)~the failure-driven prompt repair pass (App.~\ref{app:curriculum}), which adds a dedicated verification subtask to $\pi^\star$'s system prompt for trajectories of this shape.

\medskip

\subsection*{Case~3: Cost-Efficient Parallel Routing on a Multi-Statement Question}

\noindent\textit{Source:} MMLU (econometrics, item~1837).\quad
\textit{Pattern:} \textsf{oneshot} (single plan, four parallel checks plus one aggregation).\quad
\textit{Verifier:} pass.\quad
\textit{Curriculum decision:} added to $\mathcal{D}_{\text{SFT}}$ unmodified.

\smallskip
\textbf{Query.}\quad Given four statements about maximum-likelihood (ML) and ordinary-least-squares (OLS) estimation of GARCH and standard linear models: (i)~ML maximises the probability of observing the data; (ii)~GARCH is estimable only by ML; (iii)~OLS and ML coincide on slope/intercept but differ on the variance estimator; (iv)~most packages use numerical optimisation. Decide which subset is jointly true (options A-D).

\smallskip
\textbf{Trajectory.}
\begin{center}
\footnotesize
\setlength{\tabcolsep}{6pt}
\renewcommand{\arraystretch}{1.25}
\begin{tabularx}{\linewidth}{@{}p{0.20\linewidth}X@{}}
\toprule
\textbf{Schema element} & \textbf{Content} \\
\midrule
\promptkw{<plan round=1>}       & five subtasks: four parallel atomic checks $t_1,\ldots,t_4$ (one per statement) and an aggregator $t_5\prec t_1,\ldots,t_4$ that selects the option. \\
\midrule
\promptkw{<route id=t1>}        & \modelchip{gemini-2.5-flash}~~/~~\primchip{direct\_answer} \\
\promptkw{<obs t1>}             & \promptcmt{``True.''} \\
\promptkw{<route id=t2>}        & \modelchip{gemini-2.5-flash}~~/~~\primchip{direct\_answer} \\
\promptkw{<obs t2>}             & \promptcmt{``True. OLS cannot handle conditional heteroskedasticity.''} \\
\promptkw{<route id=t3>}        & \modelchip{gemini-2.5-flash}~~/~~\primchip{direct\_answer} \\
\promptkw{<obs t3>}             & \promptcmt{``True. Slope/intercept coincide under Gaussian errors; ML divides by $n$, OLS by $n{-}k$.''} \\
\promptkw{<route id=t4>}        & \modelchip{gemini-2.5-flash}~~/~~\primchip{direct\_answer} \\
\promptkw{<obs t4>}             & \promptcmt{``True.''} \\
\midrule
\promptkw{<route id=t5>}        & \modelchip{claude-opus-4-6}~~/~~\primchip{direct\_answer} \\
\promptkw{<obs t5>}             & \promptcmt{``All four statements are true; the correct option is (D).''} \\
\promptkw{<verify>}             & \statgood{pass}\quad all subtasks resolved \\
\promptkw{<final\_answer>}      & \texttt{D} \\
\bottomrule
\end{tabularx}
\end{center}

\noindent\textbf{What it illustrates.}\quad
The four sub-statements are mutually independent, so the orchestrator dispatches them in parallel through the cheapest competent worker (\modelchip{gemini-2.5-flash}, four calls) and routes only the cross-statement \emph{synthesis} step ($t_5$) to a frontier model (\modelchip{claude-opus-4-6}). The trajectory exemplifies the Pareto-efficient \textsf{oneshot} shape on which our cost-quality claims rest (\S\ref{sec:pareto}): cheap workers absorb the parallelisable atomic checks, the expensive worker is reserved for the genuinely composite decision, and dispatch latency collapses to one round of fan-out plus one aggregation rather than a five-round serial chain.


\section{Limitations}
\label{sec:limitations}

Third-party inference remains non-stationary: quotas, outages and drifting latency heterogeneously affect workers and cannot be controlled at training time. We therefore keep RL rewards centred on verifier-grounded correctness, structured delegation credit and billed token costs (\S\ref{sec:reward}) rather than wall-clock KPIs that would tether the router to fleeting provider artefacts. Architectural choices mitigate coordination overhead deterministically (\S\ref{sec:uno-orchestra}): merging planner and router into one causal backbone removes redundant hand-offs, \textsf{lazy} mode skips worker calls once a query admits a direct completion, and independent subtasks in a turn can dispatch in parallel so observations batch instead of chaining serial waits.

Residual risks remain orthogonal to API volatility. Automated verifiers are imperfect proxies for user intent: brittle schemas or flaky sandboxes distort credit despite the verifier-dominant terminal gate (\S\ref{sec:reward}). Our benchmarks, worker catalogue and tariff assumptions are moreover snapshots tied to contemporaneous endpoints; swapping models or evaluators reshapes empirical frontiers without invalidating selective delegation conceptually but does limit blanket transfer claims. Detailed hyper-parameters and per-task metrics in Apps.~\ref{app:details} through \ref{app:per-benchmark} instantiate the experiments we describe but cannot subsume unconstrained deployments.

\end{document}